\documentclass[3p]{elsarticle}
\usepackage{lineno,hyperref}
\usepackage{subcaption}
\usepackage{amsmath}
\usepackage{forest}
\usetikzlibrary{shapes,arrows,positioning,automata}

\tikzset{
  treenode/.style = {shape=rectangle, rounded corners,
                     draw, align=center,
                     top color=white,
                     bottom color=white},
  root/.style     = {treenode, font=\Large,
                     bottom color=red!30},
  env/.style      = {treenode, font=\ttfamily\normalsize},
  dummy/.style    = {circle,draw}
}
\usepackage{caption}
\usepackage{threeparttable}
\usepackage{algorithm,algpseudocode}
\usepackage[usestackEOL]{stackengine}
\usepackage{xcolor}
\edef\tmp{\the\baselineskip}
\setstackgap{L}{\tmp}
\usepackage{mathtools}

\newenvironment{codefont}{\fontfamily{lmtt}\selectfont}{\par}
\DeclareTextFontCommand{\codetext}{\codefont}

\newtheorem{hypothesis}{Hypothesis}

\journal{Journal}

\begin{document}

\begin{frontmatter}

\title{General multi-fidelity surrogate models: Framework and active learning strategies for efficient rare event simulation}

\author[A1]{Promit Chakroborty}
\author[A2]{Somayajulu L. N. Dhulipala}
\author[A2]{Yifeng Che}
\author[A2]{Wen Jiang}
\author[A2]{Benjamin W. Spencer}
\author[A2]{Jason D. Hales}
\author[A1]{Michael D. Shields}\cortext[mycorrespondingauthor]{Corresponding author; email: michael.shields@jhu.edu}

\address[A1]{Department of Civil and Systems Engineering, Johns Hopkins University, Baltimore, MD 21218, USA}
\address[A2]{Computational Mechanics and Materials, Idaho National Laboratory, Idaho Falls, ID 83415, US}


\begin{abstract}
Estimating the probability of failure for complex real-world systems using high-fidelity computational models is often prohibitively expensive, especially when the probability is small. Exploiting low-fidelity models can make this process more feasible, but merging information from multiple low-fidelity and high-fidelity models poses several challenges. 
This paper presents a robust multi-fidelity surrogate modeling strategy in which the multi-fidelity surrogate is assembled using an active learning strategy using an on-the-fly model adequacy assessment set within a subset simulation framework for efficient reliability analysis.  
The multi-fidelity surrogate is assembled by first applying a Gaussian process correction to each low-fidelity model and assigning a model probability based on the model's local predictive accuracy and cost. Three strategies are proposed to fuse these individual surrogates into an overall surrogate model based on model averaging and deterministic/stochastic model selection. The strategies also dictate which model evaluations are necessary.
No assumptions are made about the relationships between low-fidelity models, while the high-fidelity model is assumed to be the most accurate and most computationally expensive model. 
Through two analytical and two numerical case studies, including a case study evaluating the failure probability of Tristructural isotropic-coated (TRISO) nuclear fuels, the algorithm is shown to be highly accurate while drastically reducing the number of high-fidelity model calls (and hence computational cost).
\end{abstract}


\end{frontmatter}


\section{Introduction}
\label{section:Intro}

In engineering, risk assessment is a vital part of the design and analysis of complex systems, a major component of which is the estimation of failure probabilities. This is an important and challenging problem across multiple domains, such as infrastructure exposed to natural hazards (e.g., \ power grids \cite{Zio2021}, wood-frame buildings \cite{Mishra2017}), nuclear fuel safety \cite{Jiang2021}, aerospace systems reliability \cite{morio2015estimation}, and others. Consequently, it has been an ongoing topic of much research for several decades.

Reliability analysis is concerned with evaluating the probability of failure, which is given by:
\begin{equation}
\label{eqn:Pf_defn}
    P_f = \int_{\mathbf{x} \in \Omega_F} q(\mathbf{x}) dx = \int_{\mathbf{x} \in \Omega} I_{\Omega_F} (\mathbf{x}) q(\mathbf{x}) dx, 
\end{equation}
where $ \mathbf{x} $ is a random vector of the system's uncertain parameters having joint probability density $ q(\mathbf{x}) $, and $ \Omega $ is the set of all possible system states. $ \Omega_F $ represents the set of system states that correspond to failure and can be expressed as $ \Omega_F = \{ \mathbf{x}: I_{\Omega_F} (\mathbf{x})=1 \} $, where the indicator function $ I_{\Omega_F} (\mathbf{x}) $ equals $ 1 $ for states $ \mathbf{x} $ that correspond to failure and is $ 0 $ otherwise. To determine whether the system fails for a given $\mathbf{x}$, we evaluate the performance function $ g(\mathbf{x}) $, such that $ g (\mathbf{x}) \leq \mathcal{F} $ corresponds to failure, where $ \mathcal{F} $ is a pre-defined failure threshold. 
For real-world systems, the integral in Eq.~\ref{eqn:Pf_defn} is nearly always analytically intractable (due to dimensionality and/or complex failure domains). In well-behaved cases, this can be overcome through approximations, as in the case of FORM and SORM \cite{Schueller2004,ditlevsen1996structural,lemaire2013structural}. Otherwise, Monte Carlo methods \cite{Rubinstein1981MonteCarlo,Fishman1996MonteCarlo} are necessary. 
However, Monte Caro simulation is often prohibitively computationally expensive, especially for complex systems. Although many classes of innovative algorithms have been developed for improved efficiency, estimating small failure probabilities using Monte Carlo-based methods remains a challenge in general.

\subsection{Brief Review of Reliability Methods}
\label{section:Lit_Review}

One way to improve the efficiency of reliability analysis is to reduce the number of samples needed for accurate failure probability estimations (termed variance reduction). Importance Sampling (IS) is one favored approach, wherein samples are drawn from the Importance Sampling Density (ISD), an alternate density that is centered near the failure domain. Constructing an ISD requires some knowledge of the failure region. For low-to-moderate dimensionality and simple failure domains, many schemes exist for constructing the ISD, using e.g., design point-based approaches \cite{MELCHERS19893,AU1999113,DERKIUREGHIAN199837} or adaptive pre-sampling \cite{BUCHER1988119,Ang1992OptimalID}. However, for high-dimensional or complex problems, prescribing a good ISD is challenging \cite{SchuellerPradlwarter1993,au2003important}. To overcome these challenges, \cite{AuBeck2001a} introduced Subset Simulation (SuS) [detailed later]. Recently, a new class of so-called tempering methods has been introduced, which shifts the sampling distribution towards the failure region through a series of intermediate distributions \cite{Xiao2019,Catanach2018} and can be shown to generalize the SuS approach \cite{Catanach2018}.

Another means of reducing computational cost is to construct a fast-running 
data-driven surrogate model to approximate the performance function using methods such as Polynomial Chaos Expansions 
\cite{Choi2004,SudretPCKriging}, Neural Networks \cite{PAPADRAKAKIS1996145,HURTADO2001113}, or Gaussian Process Regression (GPR) \cite{Bichon2008,AKMCS}. The methods proposed by \cite{Bichon2008} and \cite{AKMCS} have been extended many times \cite{Lelievre2018a,sundar2019reliability,Razaaly2020a} and applied in conjunction with SuS \cite{AKSS}. However, these methods can break down for small failure probabilities where the failure region is far away from the nominal region of the model parameters \cite{AKEE-SS,Xu2020,SingleLF}.

The cost of individual model evaluations can also be reduced through multi-fidelity modeling, where model responses from multiple sources are incorporated. Multi-fidelity methods can be classified into three categories according to \cite{Peherstorfer2018}. \textit{Fusion} combines information from complete and accurate High Fidelity (HF) models and cheaper and less accurate Low Fidelity (LF) models using statistical tools like control variates \cite{Gorodetsky2020,pham2021ensemble} or IS \cite{Kramer2019,Peherstorfer2016} for direct estimation, or using co-Kriging to build a surrogate model from fused multi-fidelity data \cite{Yang2019,Yi2021,Zhang2022}. \textit{Filtering} evaluates the LF model first and then decides whether to evaluate the HF model. \textit{Adaptation} corrects the LF model evaluations using HF model data. Many of these multi-fidelity methods, particularly those using filtering and adaptation, are applied in an active learning framework 
\cite{Zhang2018,SingleLF}. Moreover, it is common to treat surrogate models as LF models such that methods discussed above (e.g., \ \cite{Bichon2008,AKMCS}; etc.) can be considered as filtering and adaptation-based multi-fidelity approaches. Finally, \cite{GorodetskyMFNets2020} note that many of the proposed multifidelity methods are not widely applicable because of the assumption of a fidelity hierarchy. They further proposed a generalized graph-based framework for handling all types of relationships between the available models and used it to construct both a multifidelity sampling approach \cite{GorodetskyMFNets2020} and a multifidelity surrogate modeling approach \cite{GorodetskyMFNets2021}. However, their framework requires a priori knowledge of the relationships between the models.

\subsection{Goals and Contributions of the Proposed Algorithm}
\label{section:Goals&Contributions}

The methodology proposed in this paper 
combines the multi-fidelity modeling concepts of filtering and adaptation to 
construct and actively learn GPR corrections for multiple LF models within a SuS framework for efficient reliability estimation. The method can handle LF models of any type and makes no assumptions about the relationships between the models (e.g. hierarchical or peer). Different strategies are proposed to combine the information from multiple sources, some of which do not need to evaluate each model at each sample point. The novel contributions of the work are summarized as follows:


\begin{itemize}
    \item A probabilistic framework to compare multiple LF models based on their predictive accuracy and model cost is presented. This framework is general and assumes no relationships between models.
    \item A general framework is proposed to assemble accurate surrogates from multiple LF models with associated Gaussian Process (GP) corrections, which allows for assembly strategies that do not require the evaluation of every LF model at each sample point.
    \item The proposed surrogates are applied within a subset simulation scheme to conduct reliability analysis using multiple LF models. The proposed algorithm is robust and accurate even for small failure probabilities.
    \item A novel adequacy check and model selection scheme is proposed to actively learn the surrogates ``on-the-fly'' during the subset simulation process.
    \item The proposed algorithm requires only a very small number of HF model calls, significantly improving the overall efficiency and reducing the computational cost of the analysis.
\end{itemize}

\section{Preliminaries}
\label{section:Preliminaries}

The proposed framework leverages various existing tools/methods that are widely used in reliability analysis. In particular, the framework uses a Gaussian process (GP) model for multi-fidelity modeling that is subsequently integrated into the SuS algorithm. Here, a brief review of SuS and GP modeling is first provided.  

\subsection{Subset Simulation (SuS)}
\label{section:SubSim}

\begin{figure}
\centering
\includegraphics[scale=0.25]{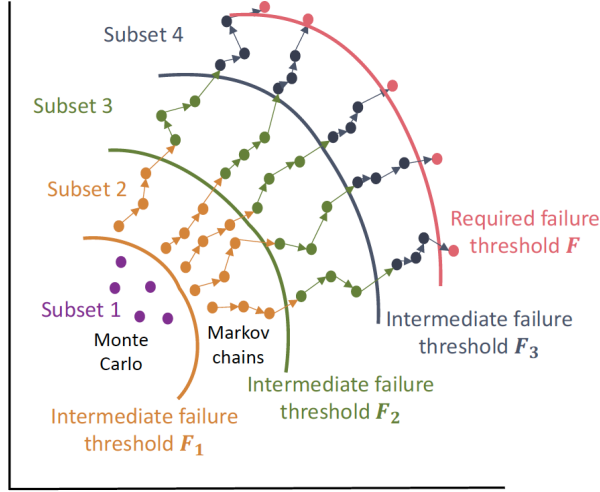}
\caption{Visualization of Subset Simulation (adapted from Figure 7 in \cite{DhulipalaGeneralReliabilityTRISO})}
\label{fig:SubSim}
\end{figure}

Subset Simulation \cite{AuBeck2001a} is a variance reduction technique that improves the efficiency of estimating small failure probabilities by representing them as the product of larger intermediate failure probabilities. This is done by constructing a series of nested subsets, $F_1\supset F_2 \supset \dots \supset F$ converging to the failure region $F$, as shown in Figure~\ref{fig:SubSim}. Each subset is sampled sequentially using a chosen Markov Chain Monte Carlo (MCMC) method, of which many have been proposed (e.g., \cite{Papaioannou2015}; \cite{ShieldsNonGaussianSubSim}), to determine $P_{s|s-1}$, which is the conditional probability that a point falling in subset $F_{s-1}$ will also fall in subset $F_s$. The failure probability is calculated as:
\begin{equation}
\label{eqn:SubSim_Pf}
    P_f = P_1 \prod_{s=2}^{N_s} P_{s|s-1},
\end{equation}
where $ N_s $ is the total number of subsets and $ P_1 $ denotes the intermediate failure probability for subset $F_1$. 
The intermediate failure domains $F_s$ are usually defined adaptively to circumvent the need for a priori knowledge of the failure surface. Instead, the intermediate failure probabilities $(P_1,~P_{2|1},~\dots,~P_{N_s-1|N_s-2})$ are fixed to a target value which (usually $0.1$). 
For further details, see \cite{AuBeck2001a}.

\subsection{Gaussian Process Regression (GPR)}
\label{section:GPR}

A Gaussian Process (GP) is a stochastic process such that every finite collection of its random variables follows a multivariate Gaussian distribution. In GPR, the target function $ g(\mathbf{x}) $ is approximated as the realization of a GP $ \mathcal{G} \left( \mathbf{x} \right) $. Thus, for a chosen covariance kernel $ K $ and given a set of training points $ \mathbf{X} = \left\{ \mathbf{x_1}, \mathbf{x_2}, \dots, \mathbf{x_n} \right\} $ with corresponding target function values $ g\left( \mathbf{X} \right) $, the value of any ``new'' point $ \mathbf{x^*} $ is estimated as $ \mathcal{G} \left( \mathbf{x^*} \right) $, such that
\begin{equation}
    \begin{bmatrix}
    g \left( \mathbf{X} \right) \\
    \mathcal{G} \left( \mathbf{x^*} \right)
    \end{bmatrix}
     \sim \mathcal{N} \left( \mathbf{0}, 
    \begin{bmatrix}
    K \left( \mathbf{X}, \mathbf{X} \right) & K \left( \mathbf{X}, \mathbf{x^*} \right) \\
    K \left( \mathbf{x^*}, \mathbf{X} \right) & K \left( \mathbf{x^*}, \mathbf{x^*} \right)
    \end{bmatrix}
    \right)
\end{equation}
where $ K \left( \mathbf{X}, \mathbf{X} \right) $ denotes the block matrix whose $ (i,j) $ \textsuperscript{th} element is $ K \left( \mathbf{x_i}, \mathbf{x_j} \right) $ such that $ \mathbf{x_i}, \mathbf{x_j} \in \mathbf{X} $, and $ K \left( \mathbf{X}, \mathbf{x^*} \right) $ and $ K \left( \mathbf{x^*}, \mathbf{X} \right) $ denote similar block matrices \cite{RasmussenWilliamsGPR}. From this, we can condition on the training point observations to show that
\begin{equation}
\label{eqn:GPR_Posterior}
\begin{aligned}
    \mathcal{G} \left( \mathbf{x^*} \right) &\sim \mathcal{N} \left( \mu_{\mathcal{G}} \left( \mathbf{x^*} \right), \sigma_{\mathcal{G}} \left( \mathbf{x^*} \right) \right) \\
    \mu_{\mathcal{G}} \left( \mathbf{x^*} \right) &= K \left( \mathbf{x^*}, \mathbf{X} \right) K \left( \mathbf{X}, \mathbf{X} \right)^{-1} g \left( \mathbf{X} \right) \\
    \sigma_{\mathcal{G}} \left( \mathbf{x^*} \right) &= K \left( \mathbf{x^*}, \mathbf{x^*} \right) - K \left( \mathbf{x^*}, \mathbf{X} \right) K \left( \mathbf{X}, \mathbf{X} \right)^{-1} K \left( \mathbf{X}, \mathbf{x^*} \right)
\end{aligned}
\end{equation}

By convention, $ \mu_{\mathcal{G}} $ is taken as the predicted response, while $ \sigma_{\mathcal{G}} $ is used to estimate the uncertainty in the prediction. The covariance kernel $K$ (having hyperparameters $\theta$) is chosen based on the requirements of the target function; the training data ($ \mathbf{X} $, $ g\left( \mathbf{X} \right) $) is used to learn the associated hyperparameters $\theta$ by minimizing the negative marginal log-likelihood $ \mathcal{L} $ given by
\begin{equation}
    \mathcal{L} \propto \frac{1}{2} \log \mid K \left( \mathbf{X}, \mathbf{X} \right) \mid + \frac{1}{2} \left( g \left( \mathbf{X} \right) \right)^T K \left( \mathbf{X}, \mathbf{X} \right) ^{-1} g \left( \mathbf{X} \right)
\end{equation}

\section{Multifidelity Modeling Strategy}
\label{section:Surrogate_Build}


The proposed framework is designed under the assumption that there exists a single HF model $ H \left( \overline{\mathbf{x}}_H \right)$ and $ N $ LF models $ L_i \left( \overline{\mathbf{x}}_{L_i} \right)$, $ i \in \left\{ 1, 2, \dots, N \right\} $. For each LF model, an independent correction $ G_i \left( \overline{\mathbf{x}} \right)$, $ i \in \left\{ 1, 2, \dots, N \right\} $ is learned using GPR, such that $ G_i $ models the difference between $ L_i $ and $ H $. Here, $ \overline{\mathbf{x}}_{L_i} $ is the vector of inputs for $L_i$, $ \overline{\mathbf{x}}_H $ is the vector of inputs for $H$, and $ \overline{\mathbf{x}} = \left\{\bigcup_{i=1}^N \overline{\mathbf{x}}_{L_i}\right\} \bigcup \overline{\mathbf{x}}_H $ denotes the vector of all inputs for all models. Note that the LF and HF models are not strictly required to take the same inputs. The $ L_i $s and $ G_i $s are used to build $ N $ ``Corrected Low-Fidelity'' models (CLFs, $ S_i \left( \overline{\mathbf{x}} \right)$) which are then probabilistically assembled into the surrogate $S \left( \overline{\mathbf{x}} \right)$, as shown in Eq.~\eqref{eqn:CLF_and_SurrogateAssembly}, by assigning individual local model probabilities $p_i(\overline{\mathbf{x}})$. (LF model evaluations are not required a priori for the computation of $p_i(\overline{\mathbf{x}})$, as described in the next section.)
\begin{equation}
\label{eqn:CLF_and_SurrogateAssembly}
    \begin{aligned}
    S_i(\overline{\mathbf{x}}) &= L_i(\overline{\mathbf{x}}_{L_i}) + G_i(\overline{\mathbf{x}}) \:, \;\; \forall i \in \left\{ 1, 2, \dots, N \right\} \\
    S(\overline{\mathbf{x}}) &= \mathcal{A}_{i=1}^N \left[ S_i(\overline{\mathbf{x}}); p_i(\overline{\mathbf{x}}) \right]
\end{aligned}
\end{equation}
where the assembly operation (i.e., $\mathcal{A}$) is defined generically in Eq.~\eqref{eqn:CLF_and_SurrogateAssembly} and can take on many forms that probabilistically combine the results from the CLFs. We specifically discuss three assembly procedures, as shown in Eq.~\eqref{eqn:Assembly_Strategies}. 
\begin{equation}
\label{eqn:Assembly_Strategies}
    S(\overline{\mathbf{x}}) = \begin{cases}
    \mathcal{A}_{i=1}^N \left[ S_i(\overline{\mathbf{x}}); p_i(\overline{\mathbf{x}}) \right] = \sum_{i=1}^N p_i(\overline{\mathbf{x}}) S_i(\overline{\mathbf{x}}) & \text{LFMA}\\
    \mathcal{A}_{i=1}^N \left[ S_i(\overline{\mathbf{x}}); p_i(\overline{\mathbf{x}}) \right] = S_k(\overline{\mathbf{x}}) \text{ s.t. } p_k(\overline{\mathbf{x}}) = \max_{i=1}^N \{p_i(\overline{\mathbf{x}})\} & \text{LFDS}\\
    \mathcal{A}_{i=1}^N \left[ S_i(\overline{\mathbf{x}}); p_i(\overline{\mathbf{x}}) \right] = S_k(\overline{\mathbf{x}}) \text{ with probability } p_k(\overline{\mathbf{x}}) & \text{LFSS}
    \end{cases}
\end{equation}
In the Low Fidelity Model Averaging (LFMA) scheme, the surrogate is the probability-weighted average of the CLFs. For Low Fidelity Deterministic Selection (LFDS), the CLF with the highest model probability is selected as the surrogate. Finally, for Low Fidelity Stochastic Selection (LFSS), one of the CLFs is chosen at random based on the assigned probabilities.

\subsection{Assigning Local Model Probabilities}
\label{section:Model_Probs}

The local model probability $ p_i(\overline{\mathbf{x}}) $ assigned to model $ i $ at the point $ \overline{\mathbf{x}} $ is the probability that model $ i $ is the ``best'' model at that point (hence local), where ``best'' is defined according to the following hypothesis:
\begin{hypothesis}
\label{hypothesis:Basic_best}
The ``best'' model is the one that has the \underline{greatest probability} of being \underline{closest} to the High-Fidelity model (i.e., it has the \underline{smallest} correction).
\end{hypothesis}

Under this hypothesis, we can derive the expression for the local model probability $ p_i(\overline{\mathbf{x}}) $ as follows. Let $ \zeta_i (\overline{\mathbf{x}}) \coloneqq \lvert G_i (\overline{\mathbf{x}}) \rvert $ denote the magnitude of the correction for model $i$ at $\overline{\mathbf{x}}$, where $G_i (\overline{\mathbf{x}})$ is the GP correction term associated with LF model $ L_i $, and let us further assume that correction terms for different models are independent. Because $G_i$ is a normally distributed random variable at a given point, i.e., \ $ G_i (\overline{\mathbf{x}}) \sim \mathcal{N} \left( \mu_i (\overline{\mathbf{x}}), \sigma_i (\overline{\mathbf{x}}) \right)$, each $ \zeta_i $ has a folded Gaussian distribution, i.e., \ $ \zeta_i (\overline{\mathbf{x}}) \sim \mathcal{N_F} \left( \mu_i (\overline{\mathbf{x}}), \sigma_i (\overline{\mathbf{x}}) \right) $.
The explicit dependence on the input point $ \overline{\mathbf{x}} $ has been dropped henceforth for brevity in notation. 

Next, let $ \Lambda $ be the set of all LF correction magnitudes, i.e., \ $ \Lambda = \left\{ \zeta_j | j \in \{1, 2, \dots N \} \right\} $, and $ \Lambda_{-i} $ be the set of all LF correction magnitudes excluding that for model $i$, i.e., \ $ \Lambda_{-i} = \left\{ \zeta_j | j \in \{1, 2, \dots i-1, i+1, \dots N \} \right\} $. Finally, let $ Z $ be the first order statistic of $ \Lambda $, i.e. $ Z = \min (\zeta_j \in \Lambda) $, and $ Z_{-i} $ be the first order statistic of $ \Lambda_{-i} $, i.e. $ Z_{-i} = \min (\zeta_j \in \Lambda_{-i}) $. The Cumulative Distribution Function (CDF) of $ Z $ is given by:
\begin{equation}
\label{eqn:CDF_Derivations}
    \begin{aligned}
        F_Z \left( z \right) &= 1 - P \left( Z > z \right) = 1 - P \left( \min_j \zeta_j > z \right) = 1 - \prod_{j=1}^N P \left( \zeta_j > z \right) \\
        &= 1 - \prod_j \left[ 1 - F_{\zeta_j} \left( z \right) \right] \text{ , } j \in \left\{ 1, 2, \dots N \right\} \\
    \end{aligned}
\end{equation}
Similarly, $F_{Z_{-i}} \left( z \right) = 1 - \prod_j \left[ 1 - F_{\zeta_j} \left( z \right) \right] \text{ , } j \in \left\{ 1, 2, \dots i-1, i+1, \dots N \right\}$. The Probability Density Function (PDF) of $Z$ is obtained through differentiation as
\begin{equation}
\label{eqn:PDF_Derivations}
    \begin{aligned}
        f_Z \left( z \right) = F_Z' \left( z \right) &= \sum_l \left[ f_{\zeta_l} \left( z \right) \prod_{j \neq l} \left\{ 1 - F_{\zeta_j} \left( z \right) \right\}\right] \text{ , } j, l \in \left\{ 1, 2, \dots N \right\} \\
    \end{aligned}
\end{equation}
Likewise, $f_{Z_{-i}} \left( z \right) = \sum_l \left[ f_{\zeta_l} \left( z \right) \prod_{j \neq l} \left\{ 1 - F_{\zeta_j} \left( z \right) \right\}\right] \text{ , } j, l \in \left\{ 1, 2, \dots i-1, i+1, \dots N \right\}$.

Next, Hypothesis 1 can be expressed mathematically by identifying the model that maximizes:
\begin{equation}
\label{eqn:Probability_Hypothesis}
    p_i = P \left( \mid G_i \mid = \min_j \mid G_j \mid \right) = P \left( \zeta_i = \min_j \zeta_j \right) = P \left( \zeta_i \leq Z_{-i} \right),
\end{equation}
which we refer to as the local model probability. Using the CDFs and PDFs derived above, we see that
\begin{equation}
    \begin{aligned}
        P \left( \zeta_i \leq Z_{-i} \right) &= \int_0^{\infty} \int_0^{z} f_{\zeta_i} \left( z_i \right) f_{Z_{-i}} \left( z \right) dz_i dz \\
        &= \int_0^{\infty} f_{Z_{-i}} \left( z \right) \left[ F_{\zeta_i}(z) - F_{\zeta_i}(0) \right] dz \\
        &= \int_0^{\infty} f_{Z_{-i}} \left( z \right) F_{\zeta_i} \left( z \right) dz \\
        &= \left[ F_{Z_{-i}} \left( z \right) F_{\zeta_i} \left( z \right) \right]_0^{\infty} - \int_0^{\infty} F_{Z_{-i}} \left( z \right) f_{\zeta_i}(z) dz \\
        &= 1 - \int_0^{\infty} \left[ 1 - \prod_{j \neq i} \left\{ 1 - F_{\zeta_j} \left( z \right) \right\} \right] f_{\zeta_i}(z) dz \\
        &= 1 - \int_0^{\infty} f_{\zeta_i}(z) dz + \int_0^{\infty} \prod_{j \neq i} \left\{ 1 - F_{\zeta_j} \left( z \right) \right\} f_{\zeta_i}(z) dz \\
    \end{aligned}
\end{equation}
This leads to the final expression for the local model probability:
\begin{equation}
\label{eqn:Model_Probability}
        p_i = P \left( \zeta_i \leq Z_{-i} \right) = \int_0^{\infty} \left[ f_{\zeta_i} (z) \prod_{j \neq i} \left\{ 1 - F_{\zeta_j} (z) \right\} \right] dz  \text{ , where } j \in \left\{ 1, 2, \dots N \right\}
\end{equation}
where $ f_{\zeta_i} (z) $ and $ F_{\zeta_i} (z) $ are the PDF and CDF of the Folded Gaussian Distribution given by:
\begin{equation}
\label{eqn:Folded_Gauss_pdf}
    f_{\zeta_i} (z) = \begin{cases}
    \frac{1}{\sigma_i \sqrt{2 \pi}} \left[ \exp \left(- \frac{\left( z - \mu_i \right)^2}{2 \sigma_i^2} \right) + \exp \left(- \frac{\left( z + \mu_i \right)^2}{2 \sigma_i^2} \right) \right] & z \in [ 0, \infty) \\
    \hfil 0 & \text{otherwise}
    \end{cases} \\
\end{equation}
\begin{equation}
\label{eqn:Folded_Gauss_cdf}
    F_{\zeta_i} (z) = \begin{cases}
    \frac{1}{2} \left[ \text{erf} \left( \frac{z - \mu_i}{\sqrt{2} \sigma_i} \right) + \text{erf} \left( \frac{z + \mu_i}{\sqrt{2} \sigma_i} \right) \right] & z \in [ 0, \infty) \\
    \hfil 0 & \text{otherwise}
    \end{cases} \\
\end{equation}
where \text{erf} is the error function given by:
\begin{equation}
    \text{erf}(z) = \frac{2}{\sqrt{\pi}} \int_0^z e^{-t^2} dt
\end{equation}

The local model probability is calculated for each model using Eq.\ \eqref{eqn:Model_Probability} at each new sample point $ \overline{\mathbf{x}} $, where the local nature of the probability, i.e., \ its dependence on the sample point $ \overline{\mathbf{x}} $, is embedded within $ \mu_i(\overline{\mathbf{x}}) $ and $ \sigma_i(\overline{\mathbf{x}}) $. Then, according to Hypothesis 1, the model that maximizes $p_i$ is the ``best'' model at point $ \overline{\mathbf{x}} $.


\subsection{Incorporating Model Cost}
\label{section:Generalized_Model_Probs}

\begin{figure}[htbp]
\centering
\includegraphics[scale=0.2]{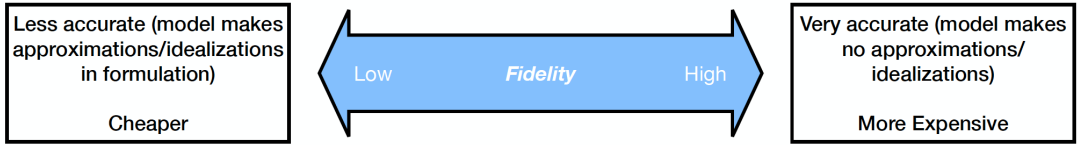}
\caption{The Fidelity Spectrum}
\label{fig: Fidelity_Spectrum}
\end{figure}

The local model probability formulation given above assumes that all LF models are peers on the fidelity spectrum (Figure~\ref{fig: Fidelity_Spectrum}). That is, no model outperforms the others over the entire input domain either in terms of computational cost or modeling accuracy, and any difference in their performance is local. However, two common cases arise where this assumption breaks down: (1) The LF models, or a subset of them, form a hierarchy in the fidelity spectrum, or (2) the LF models have widely varying costs despite having similar predictive accuracy. In the first case, any model higher on the hierarchy may be expected to outperform models lower on the hierarchy across the input domain and hence will always have a higher local model probability. In the second case, computational considerations may cause the cheaper model to be preferred. To account for these cases, the framework must be modified to consider a cost-accuracy tradeoff. This can be done by refining Hypothesis~\ref{hypothesis:Basic_best} in the following way:

\begin{hypothesis}
\label{hypothesis:Biased_best}
The ``best'' model is the one that has the \underline{greatest probability} of having the \underline{smallest} \underline{cost-biased correction}.
\end{hypothesis}

Hypothesis~\ref{hypothesis:Biased_best} can be expressed mathematically by identifying the model that maximizes:
\begin{equation}
\label{eqn:Biased_Probability_Hypothesis}
    \Tilde{p}_i = P \left( \gamma(\tau_i) \lvert G_i \rvert = \min_j \left( \gamma(\tau_j) \lvert G_j \rvert \right) \right) = P \left( \Tilde{\zeta}_i = \min_j \Tilde{\zeta}_j \right)
\end{equation}
where $\gamma(\tau_i) = \left( \tau_i \right)^{\beta}$ is a multiplicative biasing function in which $ \tau_i $ is the computational complexity of model $ i $ (usually taken to be the processor elapsed time) and $ \beta \geq 0 $ is a tuning parameter that can be used to control the biasing level, called the biasing exponent. If $ \beta > 1 $, the algorithm strictly avoids using more expensive models, and if $ \beta < 1 $, the algorithm is more permissive of using expensive models.

The cost-biased local model probability can be determined by
\begin{equation}
\label{eqn:Biased_Probability}
    \Tilde{p}_i = \int_0^{\infty} \left[ f_{\Tilde{\zeta}_i} (z) \prod_{j \neq i} \left\{ 1 - F_{\Tilde{\zeta}_j} (z) \right\} \right] dz \text{ , where } j \in \left\{ 1, 2, \dots N \right\}
\end{equation}
where $ f_{\Tilde{\zeta}_i} (z) $ and $ F_{\Tilde{\zeta}_i} (z) $ represent the PDF and CDF of the variable $\Tilde{\zeta}_i \coloneqq \gamma(\tau_i) \lvert G_i \rvert \Rightarrow \Tilde{\zeta}_i \sim \mathcal{N_F} \left( \Tilde{\mu}_i, \Tilde{\sigma}_i \right) $
having $\Tilde{\mu}_i = \gamma(\tau_i) \mu_i = \gamma(\tau_i) E \left[ G_i \right] \text{ and } \Tilde{\sigma}_i = \gamma(\tau_i) \sigma_i = \gamma(\tau_i) \sqrt{Var \left[ G_i \right]}$. Once again, the explicit dependence of $ \Tilde{p}_i $, $ \Tilde{\zeta}_i $, $ \Tilde{\mu}_i $, and $ \Tilde{\sigma}_i $ on
$ \overline{\mathbf{x}} $ has been dropped for brevity in notation; all these quantities vary from point to point within the input domain.




Note that for large values of $ \beta $, there can be stability issues in the evaluation of the probability integral depending on the numerical integration scheme used. This occurs because large multiplicative cost terms have the effect of flattening the PDFs, making the integral converge slowly. For the same reason, it is advised that the $ \tau $ values be normalized such that $ \min \tau_i = 1 $. Finally, $ \tau_i $ can be generalized to be a function of the input point $ \overline{\mathbf{x}} $; as long as $ \tau_i \geq 1 $ $ \forall i \in \left\{ 1, 2, \dots, N \right\} $ for all values of $ \overline{\mathbf{x}} $, Eq.~\ref{eqn:Biased_Probability} will hold.

\section{Active Learning to Construct Multi-Fidelity Surrogates}
\label{section:Active_Learning}

The multi-fidelity surrogates ($S(\overline{\mathbf{x}})$) described above are constructed using an active learning strategy that selects, at each iteration, new training points and the associated model(s) to run at each point. This active learning strategy is adapted from the U learning function developed by \cite{AKMCS} and integrated into SuS for reliability analysis. 

\subsection{The U Learning Function}
\label{section:U_Function}

The $ U $ Learning Function is used to adaptively select training points for a GPR surrogate for reliability analysis by balancing the exploration of regions of high predictive variance and the exploitation of regions known to be close to the failure surface. 
Given a GP surrogate $ \mathcal{G} ({\mathbf{x}}) \sim \mathcal{N} \left( \mu_{\mathcal{G}} \left( \mathbf{x} \right), \sigma_{\mathcal{G}} \left( \mathbf{x} \right) \right) $ that approximates the system response $ g ({\mathbf{x}}) $, where failure corresponded to $ g ({\mathbf{x}}) \leq 0 $, the U function is expressed as
\begin{equation}
\label{eqn:AKMCS_U_Defn}
    U \left( \mathbf{x} \right) = \frac{\lvert \mu_{\mathcal{G}} \left( \mathbf{x} \right) \rvert}{\sigma_{\mathcal{G}} \left( \mathbf{x} \right)}
\end{equation}
$ \Phi \left( - U \left( \mathbf{x} \right) \right) $, where $ \Phi ( \cdot ) $ denotes the standard normal CDF, is the probability that $ \mathcal{G} ({\mathbf{x}}) $ misclassifies the system behavior at $ \mathbf{x} $, i.e.\ the surrogate incorrectly predicts either failure or safety. 
At each iteration of the active learning, the point with the minimum $U$ value from a set of candidate samples is added to the training set, the corresponding simulation is executed, and $ \mathcal{G} ({\mathbf{x}})$ is retrained. The process terminates when $\min_i U_i>U_T$, where $U_T$ is a specified threshold above which the misclassification probability is deemed sufficiently small; typically $ U_T = 2 $ as suggested in \cite{AKMCS}.


\subsection{Adapting the U Function for Subset Simulation and Multifidelity Modeling}

Multiple works \cite{Lelievre2018a,Razaaly2020a} have shown that AK-MCS breaks down for small failure probabilities. \cite{AKSS} use a similarly trained GP surrogate within a SuS framework, which suffers from the same problems as AK-MCS \cite{AKEE-SS}. \cite{SingleLF} further showed that in a SuS framework, if the $ U $-function-based active learning step is subset-independent, breakdown occurs for low failure probabilities. Thus they propose a subset-dependent $ U $ function, which is robust for such cases.

In the present framework, the U function-based learning is further adapted for the multi-fidelity surrogates defined above. 
Here, the $ U $ value for the multi-fidelity surrogate, $S(\overline{\mathbf{x}})$, at every input point $ \overline{\mathbf{x}} $ is calculated using the subset-dependent $ U $ function introduced in \cite{SingleLF}. The $ U_s(\overline{\mathbf{x}}) $ value, which is used to check sufficiency and select training data, is calculated as follows:
\begin{equation}
\label{eqn:LFMC_U}
    U_s(\overline{\mathbf{x}}) = \frac{\lvert S\left( \overline{\mathbf{x}} \right) - \mathcal{F}_s \rvert}{\sigma \left( \overline{\mathbf{x}} \right)}
\end{equation}
where $ s $ is the subset index and $ \mathcal{F}_s $ is the corresponding failure threshold ($ \mathcal{F}_s = \mathcal{F} $ for the final subset). $ S\left( \overline{\mathbf{x}} \right) $ is the assembled surrogate model (Eqs.~\eqref{eqn:CLF_and_SurrogateAssembly},~\eqref{eqn:Assembly_Strategies}), and $ \sigma\left( \overline{\mathbf{x}} \right) $ is the associated standard deviation at $ \overline{\mathbf{x}} $, which can be calculated by noting that the surrogate at a point is a Gaussian for LFDS/LFSS, and a linear combination of Gaussians for LFMA (Eq. \eqref{eqn:Assembly_Strategies}).

\section{The Proposed Low Fidelity Model Combination Algorithm (LFMC)}
\label{section:LFMC_Algo}

\begin{figure}[htbp]
\centering
\includegraphics[width=0.75\textwidth]{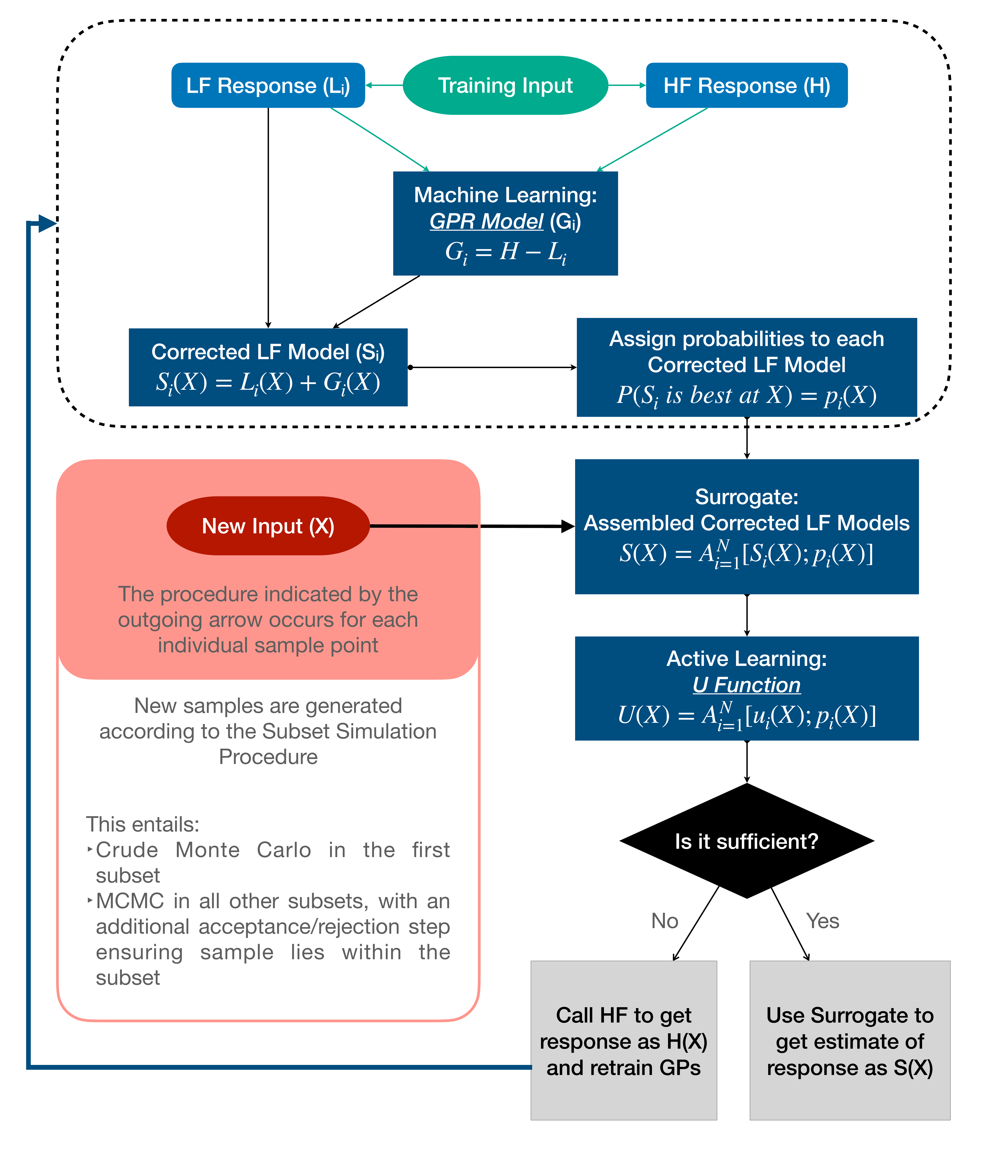}
\caption{LFMC Algorithm Flowchart}
\label{fig: Algo_Flowchart}
\end{figure}

This section collects the elements detailed above and describes a practical implementation of the proposed framework, referred to as the Low Fidelity Model Combination (LFMC)  Algorithm. Figure~\ref{fig: Algo_Flowchart} presents a flowchart of the algorithm steps, which are summarized below.

The algorithm begins with the generation of a small number ($ n_{\text{init}} $) of samples of $ \overline{\mathbf{x}}$ from the joint distribution $ q\left( \overline{\mathbf{x}} \right) $, and the evaluation of the HF model as well as each of the $ N $ LF models at each point. All the GP correction terms are trained with this initial data, i.e., $ \forall i \in \left\{ 1, 2, \dots, N \right\} $, $ G_i \left( \overline{\mathbf{x}} \right) $ is trained to predict the discrepancy $ H \left( \overline{\mathbf{x}}_H \right) - L_i \left( \overline{\mathbf{x}}_{L_i} \right) $. After this initial phase, samples are generated according to a modified multi-fidelity surrogate-based SuS framework, wherein the proposed surrogate $S( \overline{\mathbf{x}})$ is adaptively updated and used to estimate the system response at each point.

In this SuS algorithm, a total of $ N_{\text{pts}} $ sample points are generated in each subset using standard MCMC methods (using $ n_{\text{chains}} $ Markov Chains and $ n_{\text{spc}} $ samples per chain). For each new sample point $ \overline{\mathbf{x}} $, the surrogate response $ S \left( \overline{\mathbf{x}} \right) $ and learning function $ U_s \left( \overline{\mathbf{x}} \right) $ are calculated (here $ s \in \left\{ 1, 2, \dots N_s \right\}$ is the subset index, and $ N_s $ is the total number of subsets). For most points, $ S \left( \overline{\mathbf{x}} \right) $ is accepted as the system response. However, for any points where $ U_s $ indicates that the surrogate is insufficiently accurate, $ H \left( \overline{\mathbf{x}}_H \right) $ is evaluated and used as the system response instead, the GP correction is retrained, and $ S \left( \overline{\mathbf{x}} \right) $ is updated. In each subset $ s = 1, \dots, N_s - 1 $, the intermediate failure threshold $ \mathcal{F}_s $, which is necessary to evaluate $ U_s \left( \overline{\mathbf{x}} \right) $, is computed as the $ \Tilde{\pi} $\textsuperscript{th} quantile of the system responses of the samples generated within the subset thus far, where $ \Tilde{\pi} $ is the pre-selected target intermediate failure probability per \cite{AuBeck2001a}. The final subset ($ s = N_s $) is reached when the $ \Tilde{\pi} $\textsuperscript{th} quantile of the system responses $ \leq \mathcal{F} $, and here the $ U_s $ value is calculated using $ \mathcal{F}_{N_s} = \mathcal{F} $, the true failure threshold. Details for the implementation of the proposed LFMC SuS methods are provided in Algorithm \ref{Code:First_Subset} (first subset) and Algorithm \ref{Code:Intermediate_Subset} (intermediate and final subsets).

\begin{algorithm}
\caption{Sampling in the First Subset}
\label{Code:First_Subset}
\begin{codefont}
\begin{algorithmic}[1]
\Require $ N_{\text{pts}} $, $ \Tilde{\pi} $, $ q \left( \overline{\mathbf{x}} \right) $, $ H \left( \overline{\mathbf{x}}_H \right) $, $ n_{\text{chains}} $
\Ensure $ \overline{\mathbf{x}} = \left\{\bigcup_{i=1}^N \overline{\mathbf{x}}_{L_i}\right\} \bigcup \overline{\mathbf{x}}_H $
\State Create arrays SAMPLES, RESPONSES, U\_VALUES, and SEED$_1$
\For{$ l = 1 : n_{\text{chains}} $}
\For{$ m = 1 : n_{\text{spc}} $}
\State Draw $ \overline{\mathbf{x}}_{lm} \sim q \left( \overline{\mathbf{x}} \right) $ using Crude Monte Carlo
\State SAMPLES[l, m] $ \gets \overline{\mathbf{x}}_{lm} $
\State Use Algorithm~\ref{Code:Surrogate} to calculate $ S \left( \overline{\mathbf{x}}_{lm} \right) $ \\ \Comment{Treat Algorithm~\ref{Code:Surrogate} as a function.}
\State $ \mathcal{F}_1 \gets $ \textbf{QUANTILE} (RESPONSES[1:l, 1:(m-1)], $ \Tilde{\pi} $)
\State Use Algorithm~\ref{Code:Surrogate} to calculate $ U_1 \left( \overline{\mathbf{x}}_{lm} \right) $
\If{$ U_1 \left( \overline{\mathbf{x}}_{lm} \right) \geq U_T $}
\State RESPONSES[l, m] $ \gets S \left( \overline{\mathbf{x}}_{lm} \right) $, U\_VALUES[l, m] $ \gets U_1 \left( \overline{\mathbf{x}}_{lm} \right) $
\Else
\State RESPONSES[l, m] $ \gets H \left( \overline{\mathbf{x}}_{H_{lm}} \right) $, U\_VALUES[l, m] $ \gets \infty $
\State Add $ \overline{\mathbf{x}}_{lm} $ to training data
\State Retrain $ G_i $ for each $ L_i $ evaluated in the Algorithm~\ref{Code:Surrogate} call
\EndIf
\EndFor
\EndFor
\State Use SAMPLES, RESPONSES, and U\_VALUES to calculate $ P_1 $ as per equation~\ref{eqn:Subset1_Failure_Estimator}
\State Use SAMPLES, RESPONSES, and U\_VALUES to calculate $ \delta_1 $ as per equation~\ref{eqn:COV_First_Subset}
\For{$ r = 1 : n_{\text{chains}} $}
\State Select ($^1$j$_r$,$^2$j$_r$): RESPONSES[$^1$j$_r$,$^2$j$_r$] $ \leq \mathcal{F}_1 $,
\State such that ($^1$j$_r$,$^2$j$_r$) $ \neq $ ($^1$j$_t$,$^2$j$_t$), $ t \in \left\{ 1, \dots, r-1 \right \} $
\State SEED$_1$[r] $ = $ SAMPLES[$^1$j$_r$,$^2$j$_r$]
\EndFor
\end{algorithmic}
\end{codefont}
\end{algorithm}

\begin{algorithm}
\caption{Sampling in the Intermediate and Final Subsets}
\label{Code:Intermediate_Subset}
\begin{codefont}
\begin{algorithmic}[1]
\Require $ s \in \left\{ 2, 3, \dots, N_s \right\} $, $ N_{\text{pts}} $, $ \Tilde{\pi} $, $ q \left( \overline{\mathbf{x}} \right) $, $ H \left( \overline{\mathbf{x}}_H \right) $, $ n_{\text{chains}} $, $ n_{\text{spc}} $, SEED$_{s-1}$, $ \mathcal{F}_{s-1} $
\Ensure $ \overline{\mathbf{x}} = \left\{\bigcup_{i=1}^N \overline{\mathbf{x}}_{L_i}\right\} \bigcup \overline{\mathbf{x}}_H $
\State Create arrays SAMPLES, RESPONSES, U\_VALUES, and SEED$_s$
\For{$ l = 1 : n_{\text{chains}} $}
\State SAMPLES[l, 1] $ \gets $ SEED$_{s-1}$[l]
\State Copy corresponding values of RESPONSES from previous subset
\State Copy corresponding values of U\_VALUES from previous subset
\For{$ m = 1 : n_{\text{spc}} $}
\State Draw $ \overline{\mathbf{x}}_{lm} \sim q \left( \overline{\mathbf{x}} \right) $ using MCMC
\State SAMPLES[l, m] $ \gets \overline{\mathbf{x}}_{lm} $
\State Use Algorithm~\ref{Code:Surrogate} to calculate $ S \left( \overline{\mathbf{x}}_{lm} \right) $ \\ \Comment{Treat Algorithm~\ref{Code:Surrogate} as a function.}
\State $ \mathcal{F}_s \gets $ \textbf{MAX} ($ \mathcal{F} $, \textbf{QUANTILE} (RESPONSES[1:l, 1:(m-1)], $ \Tilde{\pi} $))
\State Use Algorithm~\ref{Code:Surrogate} to calculate $ U_s \left( \overline{\mathbf{x}}_m \right) $
\If{$ U_s \left( \overline{\mathbf{x}}_m \right) \geq U_T $}
\State RESPONSES[l, m] $ \gets S \left( \overline{\mathbf{x}}_{lm} \right) $,  U\_VALUES[l, m] $ \gets U_1 \left( \overline{\mathbf{x}}_{lm} \right) $
\Else
\State RESPONSES[l, m] $ \gets H \left( \overline{\mathbf{x}}_{H_{lm}} \right) $,  U\_VALUES[l, m] $ \gets \infty $
\State Add $ \overline{\mathbf{x}}_{lm} $ to training data
\State Retrain $ G_i $ for each $ L_i $ evaluated in the Algorithm~\ref{Code:Surrogate} call
\EndIf
\If{RESPONSES[l, m] $ > \mathcal{F}_{s-1} $}
\State SAMPLES[l, m] $ \gets $ SAMPLES[l, m-1]
\State RESPONSES[l, m] $ \gets $ RESPONSES[l, m-1],
\State U\_VALUES[l, m] $ \gets $ U\_VALUES[l, m-1]
\EndIf
\EndFor
\EndFor
\State Use SAMPLES, RESPONSES, and U\_VALUES to calculate $ P_{s | s-1} $ as per eq.~\ref{eqn:Intermediate_Subset_Failure_Estimator}
\State Use SAMPLES, RESPONSES, and U\_VALUES to calculate $ \delta_s $ as per eq.~\ref{eqn:COV_Intermediate_Subset}
\For{$ r = 1 : n_{\text{chains}} $ and $ s < N_s $}
\State Select ($^1$j$_r$,$^2$j$_r$): RESPONSES[$^1$j$_r$,$^2$j$_r$] $ \leq \mathcal{F}_s $,
\State such that ($^1$j$_r$,$^2$j$_r$) $ \neq $ ($^1$j$_t$,$^2$j$_t$), $ t \in \left\{ 1, \dots, r-1 \right \} $
\State SEED$_s$[r] $ = $ SAMPLES[$^1$j$_r$,$^2$j$_r$]
\EndFor
\State If $ \mathcal{F}_s = \mathcal{F} $, $ s = N_s $
\end{algorithmic}
\end{codefont}
\end{algorithm}

Some slight differences in implementation are noted when applying the LFMA, LFDS, and LFSS methods for surrogate construction (Eq.~\eqref{eqn:Assembly_Strategies}). In particular, the LFMA requires evaluation of all LF models for each surrogate evaluation, while the LFDS and LFSS schemes require only a single LF model evaluation. 
Algorithm~\ref{Code:Surrogate} codifies the steps for evaluating the surrogate model under each scheme.

\begin{algorithm}
\caption{Evaluating the Surrogate, $S( \overline{\mathbf{x}})$, and Learning Function, $ U_s(\overline{\mathbf{x}}) $}
\label{Code:Surrogate}
\begin{codefont}
\begin{algorithmic}[1]
\Require $ N $, $ L_i \left( \overline{\mathbf{x}}_{L_i} \right) $ and $ G_i \left( \overline{\mathbf{x}} \right) $ $ \forall i \in \left\{ 1, \dots N \right\} $, $ \mathcal{F}_s $, Input point $ \overline{\mathbf{x}}^* $, $ s $
\Ensure $ \overline{\mathbf{x}}^* = \left\{\bigcup_{i=1}^N \overline{\mathbf{x}}^*_{L_i}\right\} \bigcup \overline{\mathbf{x}}^*_H $
\For{$ i = 1 : N $}
\State Evaluate $ G_i \left( \overline{\mathbf{x}}^* \right) $ and $ \sigma_i \left( \overline{\mathbf{x}}^* \right) $
\State Evaluate $ \Tilde{p}_i \left( \overline{\mathbf{x}}^* \right) $ using equation~\ref{eqn:Biased_Probability}
\EndFor
\If{Assembly Strategy is LFMA}
\State $\mathcal{V} \left( \overline{\mathbf{x}}^* \right) = 0$
\For{$ i = 1 : N $}
\State Evaluate $ L_i \left( \overline{\mathbf{x}}_{L_i}^* \right) $
\State $ S_i \left( \overline{\mathbf{x}}^* \right) = L_i \left( \overline{\mathbf{x}}_{L_i}^* \right) +  G_i \left( \overline{\mathbf{x}}^* \right) $ \Comment{Equation~\ref{eqn:CLF_and_SurrogateAssembly}}
\State $\mathcal{V} \left( \overline{\mathbf{x}}^* \right) = \mathcal{V} \left( \overline{\mathbf{x}}^* \right) + \left( \Tilde{p}_i \left( \overline{\mathbf{x}}^* \right) \sigma_i \left( \overline{\mathbf{x}}^* \right) \right)^2$
\EndFor
\State $ S \left( \overline{\mathbf{x}}^* \right) = \sum_{i=1}^N \Tilde{p}_i \left( \overline{\mathbf{x}}^* \right) S_i \left( \overline{\mathbf{x}}^* \right) $ \Comment{Equation~\ref{eqn:Assembly_Strategies}}
\State $\sigma \left( \overline{\mathbf{x}}^* \right) = \sqrt{\mathcal{V} \left( \overline{\mathbf{x}}^* \right)}$
\State $ U_s \left( \overline{\mathbf{x}}^* \right) = \frac{\mid S \left( \overline{\mathbf{x}}^* \right) - \mathcal{F}_s \mid}{\sigma \left( \overline{\mathbf{x}}^* \right) }$
\Comment{Equation~\ref{eqn:LFMC_U}}
\ElsIf{Assembly Strategy is LFDS}
\State Select $ k $: $ \Tilde{p}_k \left( \overline{\mathbf{x}}^* \right) = \max_{i=1}^N \left\{ \Tilde{p}_i \left( \overline{\mathbf{x}}^* \right) \right\} $ \Comment{Equation~\ref{eqn:Assembly_Strategies}}
\State $ S \left( \overline{\mathbf{x}}^* \right) = S_k \left( \overline{\mathbf{x}}^* \right) = L_k \left( \overline{\mathbf{x}}_{L_k}^* \right) +  G_k \left( \overline{\mathbf{x}}^* \right) $ \Comment{Equation~\ref{eqn:CLF_and_SurrogateAssembly},~\ref{eqn:Assembly_Strategies}}
\State $ U_s \left( \overline{\mathbf{x}}^* \right) = \frac{\mid S_k \left( \overline{\mathbf{x}}^* \right) - \mathcal{F}_s \mid}{\sigma_k \left( \overline{\mathbf{x}}^* \right) }$ \Comment{Equation~\ref{eqn:LFMC_U}}
\ElsIf{Assembly Strategy is LFSS}
\State Select $ k $ with probability $ \Tilde{p}_k \left( \overline{\mathbf{x}}^* \right) $ \Comment{Equation~\ref{eqn:Assembly_Strategies}}
\State $ S \left( \overline{\mathbf{x}}^* \right) = S_k \left( \overline{\mathbf{x}}^* \right) = L_k \left( \overline{\mathbf{x}}_{L_k}^* \right) +  G_k \left( \overline{\mathbf{x}}^* \right) $ \Comment{Equation~\ref{eqn:CLF_and_SurrogateAssembly},~\ref{eqn:Assembly_Strategies}}
\State $ U_s \left( \overline{\mathbf{x}}^* \right) = \frac{\mid S_k \left( \overline{\mathbf{x}}^* \right) - \mathcal{F}_s \mid}{\sigma_k \left( \overline{\mathbf{x}}^* \right) }$ \Comment{Equation~\ref{eqn:LFMC_U}}
\EndIf
\State \textbf{RETURN} $ S \left( \overline{\mathbf{x}}^* \right) $, $ U_s \left( \overline{\mathbf{x}}^* \right) $, all evaluated $ L_i \left( \overline{\mathbf{x}}_{L_i}^* \right) $,
\State $ \text{ }\text{ }\text{ }\text{ }\text{ }\text{ }\text{ } k $ if Assembly Strategy is LFDS or LFSS
\end{algorithmic}
\end{codefont}
\end{algorithm}

\section{\textbf{Estimators for Reliability Analysis}}
\label{section:Estimators}


Per the Subset Simulation method, the probability of failure can be estimated through the product of probabilities in Eq.~\eqref{eqn:SubSim_Pf}. However, the intermediate failure probabilities require the following special estimators
\begin{align}
    P_1 &= \frac{1}{N_{\text{pts}}} \sum_{l = 1}^{n_{\text{chains}}} \sum_{m = 1}^{n_{\text{spc}}} \mathcal{P}_{lm}^{\left( 1 \right)} \label{eqn:Subset1_Failure_Estimator}\\
    P_{s|s-1} &= \frac{1}{N_{\text{pts}}} \sum_{l = 1}^{n_{\text{chains}}} \sum_{m = 1}^{n_{\text{spc}}} \mathcal{P}_{lm}^{\left( s \right)} \text{ , } \forall s = \left\{ 2, 3, \dots, N_s \right\} \label{eqn:Intermediate_Subset_Failure_Estimator}
\end{align}
where $ n_{\text{spc}} = \frac{N_{\text{pts}}}{n_{\text{chains}}} $. The value of $ \mathcal{P}_{lm}^{\left( s \right)} $, termed the probability of point failure, is calculated for each point in each subset using the Law of Total Probability as
\begin{multline}
\label{eqn:LoTP_for_sample_point_failure}
    \mathcal{P}_{lm}^{\left( s \right)} = P \left( I_{\mathcal{F}_s} \left( \overline{\mathbf{x}}_{lm} \right) = 1 \mid I_{\mathcal{F}_s, S} \left( \overline{\mathbf{x}}_{lm} \right) = 1 \right) P \left( I_{\mathcal{F}_s, S} \left( \overline{\mathbf{x}}_{lm} \right) = 1 \right) \\
    + P \left( I_{\mathcal{F}_s} \left( \overline{\mathbf{x}}_{lm} \right) = 1 \mid I_{\mathcal{F}_s, S} \left( \overline{\mathbf{x}}_{lm} \right) = 0 \right) P \left( I_{\mathcal{F}_s, S} \left( \overline{\mathbf{x}}_{lm} \right) = 0 \right)
\end{multline}
where $ \overline{\mathbf{x}}_{lm} $ is the $ m^{\text{th}} $ sample point in the $ l^{\text{th}} $ chain in subset $ s $ ($ \forall s = \left\{ 1, 2, \dots, N_s \right\} $), $ I_{\mathcal{F}_s, S} \left( \overline{\mathbf{x}}_{lm} \right) $ indicates whether the surrogate predicts failure at point $ \overline{\mathbf{x}}_{lm} $, and $ I_{\mathcal{F}_s} \left( \overline{\mathbf{x}}_{lm} \right) $ indicates whether the ``true'' (i.e. HF model) response indicates failure at $ \overline{\mathbf{x}}_{lm} $. Note that the subscript $ \mathcal{F}_s $ specifies that the failure indicated is with respect to the intermediate failure threshold for the corresponding subset. More rigorously, these indicator functions can be defined as
\begin{align}
    I_{\mathcal{F}_s, S} \left( \overline{\mathbf{x}}_{lm} \right) &= 
    \begin{cases}
    1 & \text{if } S \left( \overline{\mathbf{x}}_{lm} \right) \leq \mathcal{F}_s \\
    0 & \text{otherwise}
    \end{cases} \\
    I_{\mathcal{F}_s} \left( \overline{\mathbf{x}}_{lm} \right) &= 
    \begin{cases}
    1 & \text{if HF response at } \overline{\mathbf{x}}_{lm} \leq \mathcal{F}_s \\
    0 & \text{otherwise}
    \end{cases}
\end{align}

The terms in Eq.~\eqref{eqn:LoTP_for_sample_point_failure} have intuitive interpretations. $ P \left( I_{\mathcal{F}_s, S} \left( \overline{\mathbf{x}}_{lm} \right) = 1 \right) $ and $ P \left( I_{\mathcal{F}_s, S} \left( \overline{\mathbf{x}}_{lm} \right) = 0 \right) $ are the probabilities that the surrogate model predicts either failure or safety of the system, respectively. $ P \left( I_{\mathcal{F}_s} \left( \overline{\mathbf{x}}_{lm} \right) = 1 \mid I_{\mathcal{F}_s, S} \left( \overline{\mathbf{x}}_{lm} \right) = 0 \right) $ is the probability of misclassification of safety by the surrogate, and $ P \left( I_{\mathcal{F}_s} \left( \overline{\mathbf{x}}_{lm} \right) = 1 \mid I_{\mathcal{F}_s, S} \left( \overline{\mathbf{x}}_{lm} \right) = 1 \right) $ is the probability that the surrogate correctly predicts failure.

Eq.~\eqref{eqn:LoTP_for_sample_point_failure} can be simplified further. First, notice that $ P \left( I_{\mathcal{F}_s, S} \left( \overline{\mathbf{x}}_{lm} \right) \right) = I_{\mathcal{F}_s, S} \left( \overline{\mathbf{x}}_{lm} \right) $ because the surrogate deterministically predicts safety or failure based on the estimated response and the failure threshold. Next, recall that $ \Phi \left( - U \left( \mathbf{x} \right) \right) $, represents the probability of misclassification. 
Substituting this relation for the conditional probabilities in Eq.~\eqref{eqn:LoTP_for_sample_point_failure}, we can represent $ \mathcal{P}_{lm}^{\left( s \right)} $ simply as
\begin{equation}
\label{eqn:Simplified_sample_point_failure}
    \mathcal{P}_{lm}^{\left( s \right)} = 
    \begin{cases}
    \Phi \left( U_s \left( \overline{\mathbf{x}}_{lm} \right) \right) & \text{if } I_{\mathcal{F}_s, S} \left( \overline{\mathbf{x}}_{lm} \right) = 1 \\
    \Phi \left( - U_s \left( \overline{\mathbf{x}}_{lm} \right) \right) & \text{if } I_{\mathcal{F}_s, S} \left( \overline{\mathbf{x}}_{lm} \right) = 0
    \end{cases}
\end{equation}


To derive the Coefficient of Variation (COV) estimator, we follow the steps from \cite{AuBeck2001a}, replacing their indicator of failure of a sample point with our probability of point failure $ \mathcal{P}_{lm} $ from Eq.~\eqref{eqn:Simplified_sample_point_failure}. 
As in \cite{AuBeck2001a}, the final COV estimate is decomposed into COV estimates for each of the intermediate subsets as
\begin{equation}
\label{eqn:COV_Expression}
    \delta_f = \sqrt{\sum_{s=1}^{N_s} \delta_s^2 }
\end{equation}
where 
\begin{equation}
\label{eqn:COV_First_Subset}
    \delta_1 = \sqrt{\frac{1-P_1}{P_1 N_{\text{pts}}}}
\end{equation}
for the first subset. To compute $ \delta_s $, $ 2 \leq s \leq N_s $, we use the following set of equations
\begin{align}
    \delta_s &= \sqrt{ \frac{\left( 1 - P_{s|s-1} \right)}{P_{s|s-1} N_{\text{pts}}} \left[ 1 + \gamma_s \right] }
    \label{eqn:COV_Intermediate_Subset} \\
    \gamma_s &= 2 \sum_{\lambda=1}^{n_{\text{spc}} - 1} \left( 1 - \frac{\lambda }{n_{\text{spc}}} \right) \rho_s (\lambda)
    \label{eqn:COV_Gamma} \\
    \rho_s (\lambda) &= \frac{R_s (\lambda)}{R_s (0)}
    \label{eqn:COV_Rho} \\
    R_s (\lambda) &= \left[ \frac{1}{N_{\text{pts}} - \lambda n_{\text{chains}}} \sum_{l = 1}^{n_{\text{chains}}} \sum_{m = 1}^{n_{\text{spc}}-\lambda} \mathcal{P}_{lm}^{\left( s \right)} \mathcal{P}_{l \left(m + \lambda \right)}^{\left( s \right)} \right] - P_{s|s-1}^2
    \label{eqn:COV_Autocorrelation_with_lag} \\
    R_s (0) &= Var \left[ \mathcal{P}_{lm}^{\left( s \right)} \right]
    \label{eqn:COV_Var_of_point_failure}
\end{align}

Note from Eq.~\eqref{eqn:Simplified_sample_point_failure} that $ \mathcal{P}_{lm}^{\left( s \right)} $ is a two-point random variable. However, with the retraining of the surrogate based on the sufficiency criterion, we ensure that $ U_s \left( \overline{\mathbf{x}}_{lm} \right) \geq U_T $. Therefore, in practice, $ \Phi \left( U_s \left( \overline{\mathbf{x}}_{lm} \right) \right) \approx 1 $ and $ \Phi \left( -U_s \left( \overline{\mathbf{x}}_{lm} \right) \right) \approx 0 $. So we approximate $ \mathcal{P}_{lm}^{\left( s \right)} $ as a Bernoulli random variable to compute
\begin{equation}
\label{eqn:COV_Var_of_point_failure_simplified}
    R_s (0) = P_{s|s-1} \left( 1 - P_{s|s-1} \right)
\end{equation}

\section{Case Studies and Numerical Validation}

In this section, we apply the LFMC algorithm to four case studies. We compare its performance with standard Monte Carlo where feasible and standard subset simulation without the use of a surrogate and provide some insights into the algorithm. 
First, the algorithm is applied to two benchmark analytical examples to highlight the ability of the algorithm to learn complex failure domains. Then, we present the algorithm's performance for two numerical case studies, one of moderate complexity and one of high complexity. 

\subsection{Analytical Case Studies}

In this section, we apply the proposed LFMC algorithm for two analytical functions where the LF model is derived from the full model form. We note that, for these examples, the cost of the LF and HF models are comparable (and cheap) so we do not incorporate model cost into the selection algorithm.

\subsubsection{Four-Branch Function}

The four-branch function is a widely used benchmark problem in reliability analysis defined by the following equation
\begin{equation}
\label{eqn:Four-Branch_Definition}
    F \left( \mathbf{x} \right) = \min
    \begin{cases}
    3 + \frac{\left( x_1 - x_2 \right)^2}{10} - \frac{x_1 + x_2}{\sqrt{2}} \\
    3 + \frac{\left( x_1 - x_2 \right)^2}{10} + \frac{x_1 + x_2}{\sqrt{2}} \\
    x_1 - x_2 + \frac{6}{\sqrt{2}} \\
    x_2 - x_1 + \frac{6}{\sqrt{2}}
    \end{cases}
\end{equation}
where the input $ \mathbf{x} = \left\{ x_1, x_2 \right\} $ is a two-dimensional uncorrelated standard normal random vector. We run the LFMC framework using the four-branch function as the HF model and each of the four branches as individual LF models, defined as
\begin{gather}
\label{eqn:Four-Branch_LFMC_Models_Definition}
    L_1 \left( \mathbf{x} \right) = 3 + \frac{\left( x_1 - x_2 \right)^2}{10} - \frac{x_1 + x_2}{\sqrt{2}} \\
    L_2 \left( \mathbf{x} \right) = 3 + \frac{\left( x_1 - x_2 \right)^2}{10} + \frac{x_1 + x_2}{\sqrt{2}} \\
    L_3 \left( \mathbf{x} \right) = x_1 - x_2 + \frac{6}{\sqrt{2}} \\
    L_4 \left( \mathbf{x} \right) = x_2 - x_1 + \frac{6}{\sqrt{2}} 
\end{gather}
Note that for this example, $ \overline{\mathbf{x}} =  \overline{\mathbf{x}}_{L_1} = \overline{\mathbf{x}}_{L_2} = \overline{\mathbf{x}}_{L_3} = \overline{\mathbf{x}}_{L_4} = \overline{\mathbf{x}}_H = \mathbf{x} $. 
Failure corresponds to $ H \left( \mathbf{x} \right) = F \left( \mathbf{x} \right) \leq 0$. 

\begin{figure}[!htbp]
\centering
\begin{subfigure}{.49\textwidth}
  \centering
  \includegraphics[width=\linewidth]{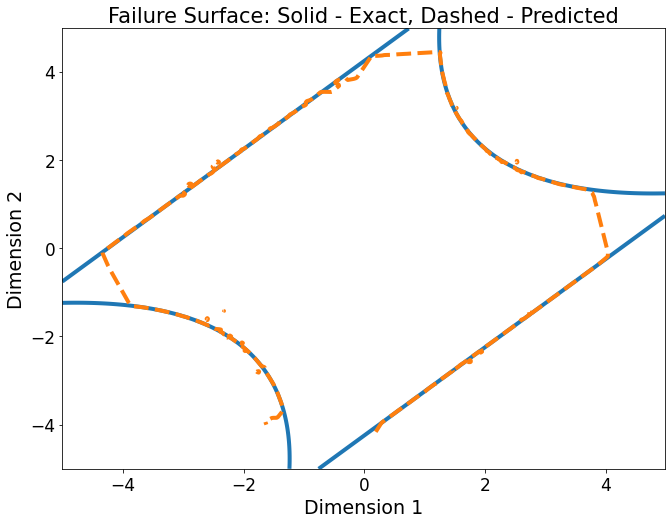}
  \caption{}
  \label{fig:Four-Branch Failure Boundaries}
\end{subfigure}%
\begin{subfigure}{.49\textwidth}
  \centering
  \includegraphics[width=\linewidth]{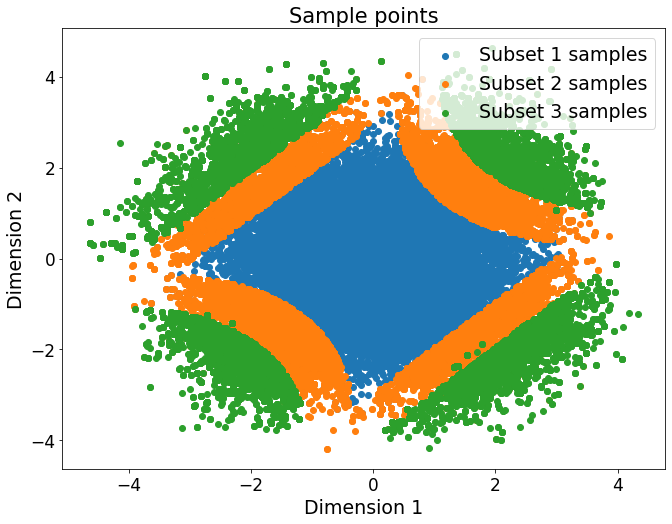}
  \caption{}
  \label{fig:Four-Branch Samples}
\end{subfigure}
\begin{subfigure}{.49\textwidth}
  \centering
  \includegraphics[width=\linewidth]{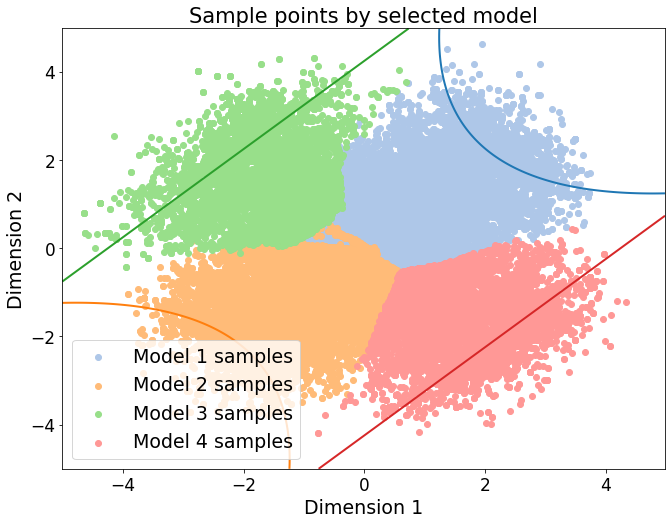}
  \caption{}
  \label{fig:Four-Branch Samples by Model Call}
\end{subfigure}%
\begin{subfigure}{.49\textwidth}
  \centering
  \includegraphics[width=\linewidth]{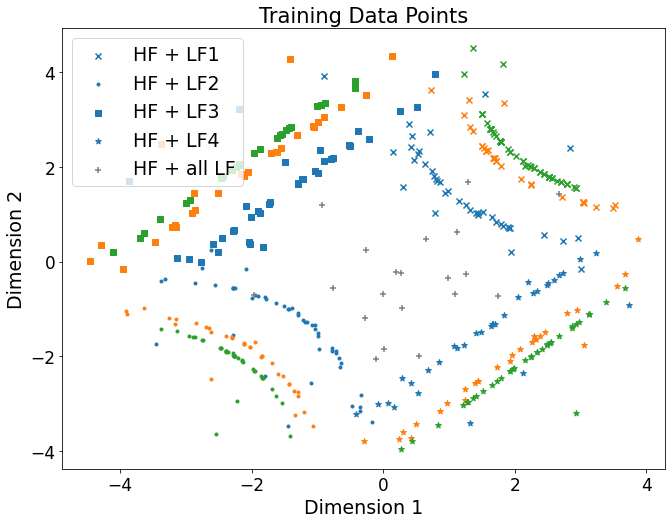}
  \caption{}
  \label{fig:Four-Branch Training Samples}
\end{subfigure}
\caption{Four Branch Function (a) Comparison between exact failure boundary and that predicted by the LFDS surrogate. (b) Samples drawn by LFMC - LFDS colored by subset. (c) Samples drawn by LFMC - LFDS colored by LF model call. Initial training set is removed. The contour of the failure surface is added, with each branch color-coded to the corresponding LF model. (d) Location of surrogate training points (HF +LF model evaluations) for LFMC - LFDS. Colors follow the same scheme as (b).}
\label{fig:Four-Branch Function Results}
\end{figure}

The performance of the LFMC algorithm is presented in Table~\ref{tab:Four-Branch_Results} and illustrated in Figure~\ref{fig:Four-Branch Function Results}, for $ 20,000 $ samples per subset and an initial set of $ n_{\text{init}} = 20 $ randomly generated training samples. At these training points, all LF models and the HF model are evaluated and the initial surrogate models are assembled.
\begin{table}[h]
    \centering
    \begin{tabular}{c c c c}
    \hline \hline
      Algorithm & Estimated $ P_f $ (COV) & \# HF calls (\% of total) & $ R^2 $ value \\
      \hline
      Crude Monte Carlo & 4.45E-3 (0.015) & 1 million (100\%) & - \\
      Subset Simulation & 4.37E-3 (0.047) & 60000 (100\%) & - \\
      LFMC - LFDS & 4.58E-3 (0.052) & 470 (0.78\%) & 0.9905 \\
      LFMC - LFSS & 3.97E-3 (0.053) & 599 (0.99\%) & 0.9555 \\
      LFMC - LFMA & 5.16E-3 (0.052) & 588 (0.98\%) & 0.9842 \\
    \hline
    \end{tabular}
    \caption{Results for the Four-Branch Function (60,000 total samples generated for LFMC methods)}
    \label{tab:Four-Branch_Results}
\end{table}

Table~\ref{tab:Four-Branch_Results} clearly shows that the LFMC framework predicts accurate failure probabilities with only a tiny fraction of the HF model calls compared to Crude Monte Carlo and standard Subset Simulation. We also note that the LFDS assembly strategy performs the best in terms of both the number of HF model calls and $ R^2 $ fit of the surrogate, which is further illustrated by plotting the surrogate limit surface in Figure~\ref{fig:Four-Branch Failure Boundaries}. Figure~\ref{fig:Four-Branch Samples} further shows the samples drawn in each subset using LFDS. We emphasize that the LFDS strategy only uses the ``best'' LF model at each point. Consequently, the surrogate predicts the response using the LF model that corresponds to the relevant branch of the HF model at that point. This is clearly illustrated in Figure~\ref{fig:Four-Branch Samples by Model Call}, which shows all the samples generated by the algorithm, colored by the LF model used to estimate the response at that point. Clearly, the surrogate is calling the LF associated with the nearest branch. 

Figure~\ref{fig:Four-Branch Training Samples}, which shows the surrogate model training points by LF model call, also follows this pattern. It clearly shows how the surrogate training points for each LF model are restricted to the region where that LF model is accurate, which enhances performance by ensuring that training samples are not wasted to train a surrogate model in regions where its LF model is inaccurate. This explainability 
is an added benefit of the LFMC framework. LFMC not only accurately and efficiently estimates failure probabilities but also provides insights into each LF model's strengths (and limitations).


\subsubsection{Rastrigin Function}

The Rastrigin function is a benchmark problem used to test whether reliability methods can resolve complex failure boundaries. It is defined as follows
\begin{equation}
    \label{eqn:Rastrigin_Definition}
    F \left( \mathbf{x} \right) = 10 - \sum_{i=1}^2 \left( x_i^2 - 5 \cos{2 \pi x_i} \right)
\end{equation}
where the input $ \mathbf{x} = \left\{ x_1, x_2 \right\} $ is a two-dimensional uncorrelated standard normal random vector. Here, we define two sets of LF models, which are called Type 1 and Type 2, to explore the impact of using different LF model definitions on the algorithm's performance. The model definitions for Type 1 are
\begin{gather}
\label{eqn:Rastrigin_Type1_LFMC_Models_Definition}
    L_1^{(1)} \left( \mathbf{x} \right) = 10 - \left( x_1^2 - 5 \cos{2 \pi x_1} \right) \\
    L_2^{(1)} \left( \mathbf{x} \right) = 10 - \left( x_2^2 - 5 \cos{2 \pi x_2} \right) 
\end{gather}
and those for Type 2 are
\begin{gather}
    L_1^{(2)} \left( \mathbf{x} \right) = 10 - \left( x_1^2 + x_2^2 \right)\label{eqn:Rastrigin_Type_2_Quadratic} \\
    L_2^{(2)} \left( \mathbf{x} \right) = 10 + \left( 5 \cos{2 \pi x_1} + 5 \cos{2 \pi x_2} \right) \label{eqn:Rastrigin_Type_2_Cosine} 
\end{gather}
Note that $ \overline{\mathbf{x}} =  \overline{\mathbf{x}}_{L_1^{(2)}} = \overline{\mathbf{x}}_{L_2^{(2)}} = \overline{\mathbf{x}}_H = \mathbf{x} $, while $ \overline{\mathbf{x}}_{L_i^{(1)}} = x_i $. 
Failure corresponds to $ H \left( \mathbf{x} \right) = F \left( \mathbf{x} \right) \leq 0 $. We see that Type 1 and Type 2 LF models correspond to different groupings of terms in the HF model, where in Type 1, the summation is simply split to form 2 models, and in Type 2, the quadratic and cosine terms are grouped to form the LF models.

Table~\ref{tab:Rastrigin_Results} presents the performance of the LFMC algorithm for both sets of LF models, with $ n_{\text{init}} = 20 $ and $ 30,000 $ samples per subset in all cases.
We again explore the LFDS strategy in more detail here and discuss the other strategies in the next section. 
\begin{table}[!ht]
    \centering
    \begin{tabular}{c c c c}
    \hline \hline
      Algorithm & Estimated $ P_f $ (COV) & \# HF calls (\% of total) & $ R^2 $ value \\
      \hline
      Crude Monte Carlo & 7.31E-2 (0.003) & 1 million (100\%) & - \\
      Subset Simulation & 7.23E-2 (0.016) & 80000 (100\%) & - \\
      Type 1: LFMC - LFDS & 7.39E-2 (0.019) & 108 (0.18\%) & 0.9998 \\
      Type 1: LFMC - LFSS & 7.43E-2 (0.019) & 118 (0.19\%) & 0.9999 \\
      Type 1: LFMC - LFMA & 7.10E-2 (0.019) & 90 (0.15\%) & 0.9999 \\
      Type 2: LFMC - LFDS & 7.14E-2 (0.019) & 460 (0.77\%) & 0.9968 \\
      Type 2: LFMC - LFSS & 7.22E-2 (0.019) & 474 (0.79\%) & 0.9987 \\
      Type 2: LFMC - LFMA & 7.38E-2 (0.019) & 446 (0.74\%) & 0.9973 \\
      Type 2: Single LF: Quadratic & 7.39E-2 (0.019) & 506 (0.84\%) & 0.9979 \\
      Type 2: Single LF: Cosine & 7.06E-2 (0.019) & 75 (0.13\%) & 0.9999 \\
    \hline
    \end{tabular}
    \caption{Results for the Rastrigin Function (60,000 total samples generated for LFMC methods)}
    \label{tab:Rastrigin_Results}
\end{table}

We see that the failure probability is predicted accurately for both Type 1 and Type 2 LF models. However, the number of HF calls required for the two cases is vastly different. 
This suggests that the algorithm's performance depends on the LF models provided, even for the same HF model. This makes sense because, for different LF models, the function that the GP correction term needs to learn will be different. If this function is more complex, more HF calls will be required. Rastrigin Type 2 highlights the reduction in performance when a poor set of LF models is provided. The quadratic LF model (Eq.~\eqref{eqn:Rastrigin_Type_2_Quadratic}) ignores the oscillatory behavior of the Rastrigin function, and therefore the corresponding correction GP requires a large amount of training data. 

In Table~\ref{tab:Rastrigin_Results} 
the LFMC algorithm is further compared to the single LF model algorithm from \cite{SingleLF}.
Here, the two LF models of Type 2 are used in turn as the single LF model, and the results are presented as the cases ``Single LF: Quadratic'' and ``Single LF: Cosine''. 
We see that the single LF case with the cosine LF model alone performs significantly better than all other cases for Type 2. This highlights the fact that using multiple LF models is not always more efficient than using only a single LF model. 
We note, however, that the LFMC algorithm is not necessarily designed to be more efficient than the single LF model case. Instead, the hypothesis underpinning its development 
prioritizes the selection of LF models that \textit{locally} best compare to the HF model. This may not always yield the surrogate whose correction can be learned with minimal training data -- as is the case for the single LF cosine model whose correction is a simple quadratic function and is therefore easy to learn from a few training points. However, if multiple LF models are available, we cannot usually know \textit{a priori} if a single LF model will be ``easy'' to correct and, as evidenced by the single quadratic LF case, selection of a single LF that is difficult to correct will lead to inefficiency. 
Instead, our probability assignment scheme provides a robust, general, and rigorous way to handle information from multiple LF models in the absence of such insights.

\begin{figure}[!htbp]
\centering
\begin{subfigure}{.49\textwidth}
  \centering
  \includegraphics[width=\linewidth]{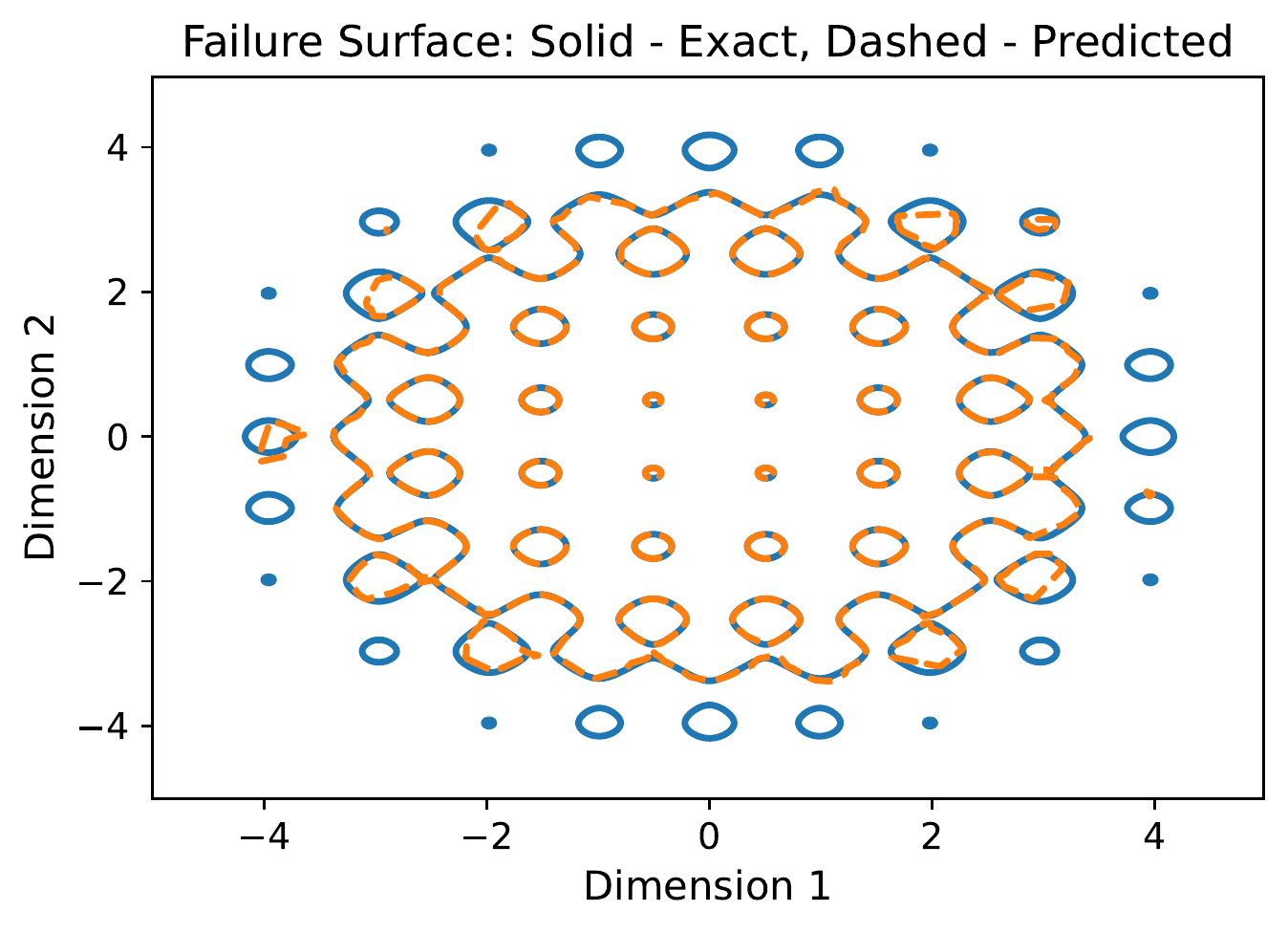}
  \caption{}
  \label{fig:Rastrigin Type 1 Failure Boundaries}
\end{subfigure}%
\begin{subfigure}{.49\textwidth}
  \centering
  \includegraphics[width=\linewidth]{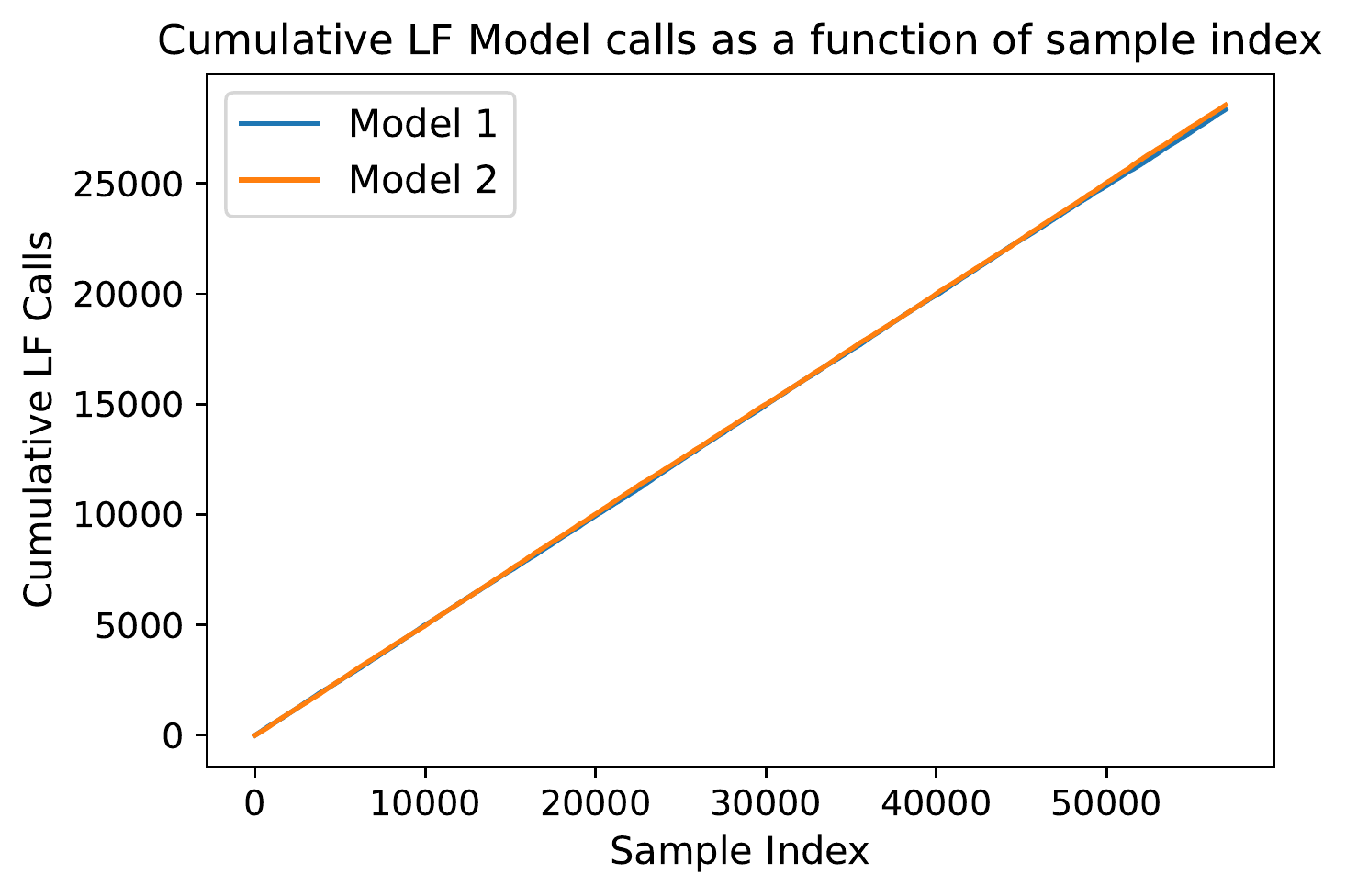}
  \caption{}
  \label{fig:Rastrigin Type 1 Cumulative LF}
\end{subfigure}
\begin{subfigure}{.49\textwidth}
  \centering
  \includegraphics[width=\linewidth]{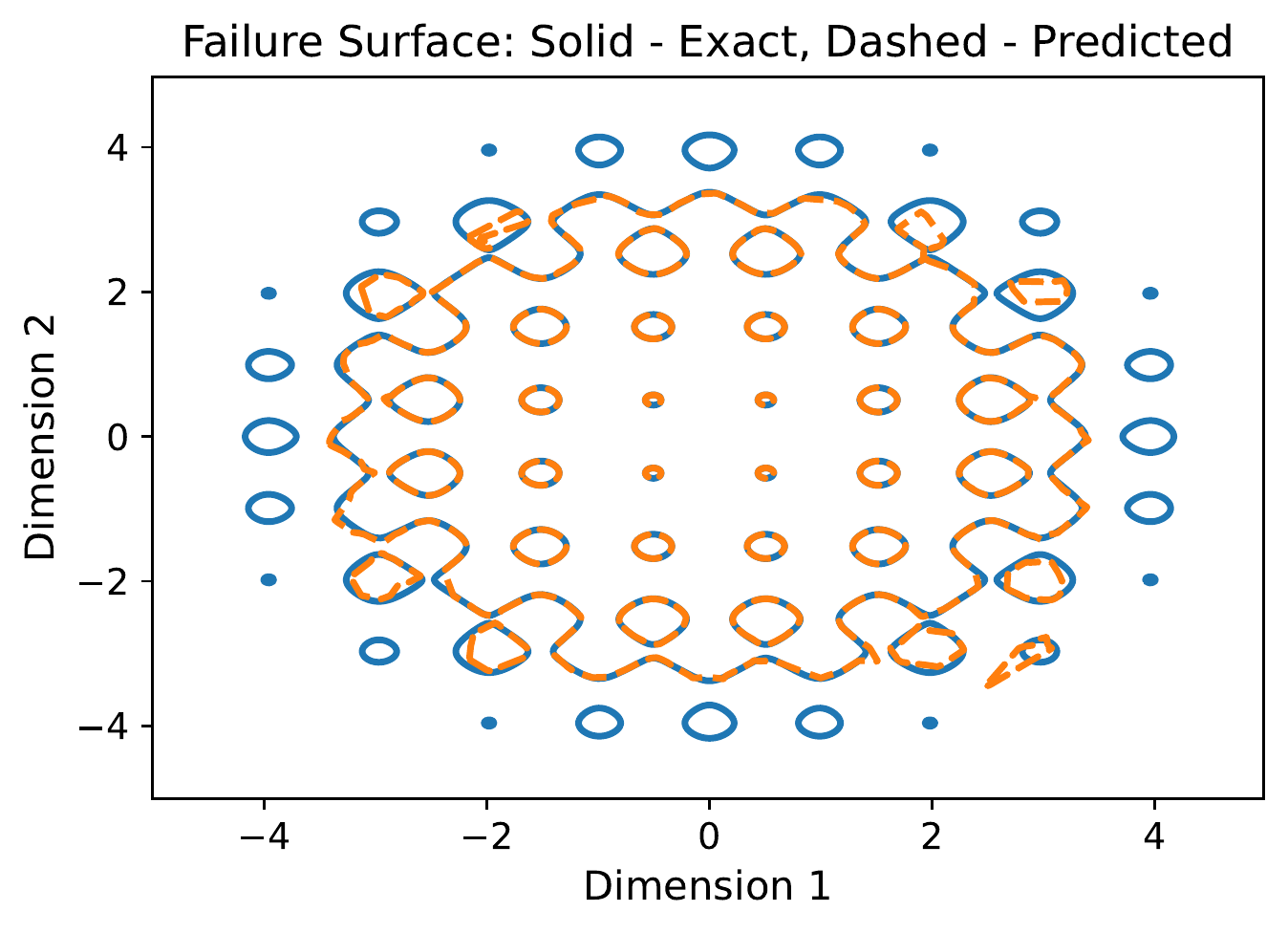}
  \caption{}
  \label{fig:Rastrigin Type 2 Failure Boundaries}
\end{subfigure}%
\begin{subfigure}{.49\textwidth}
  \centering
  \includegraphics[width=\linewidth]{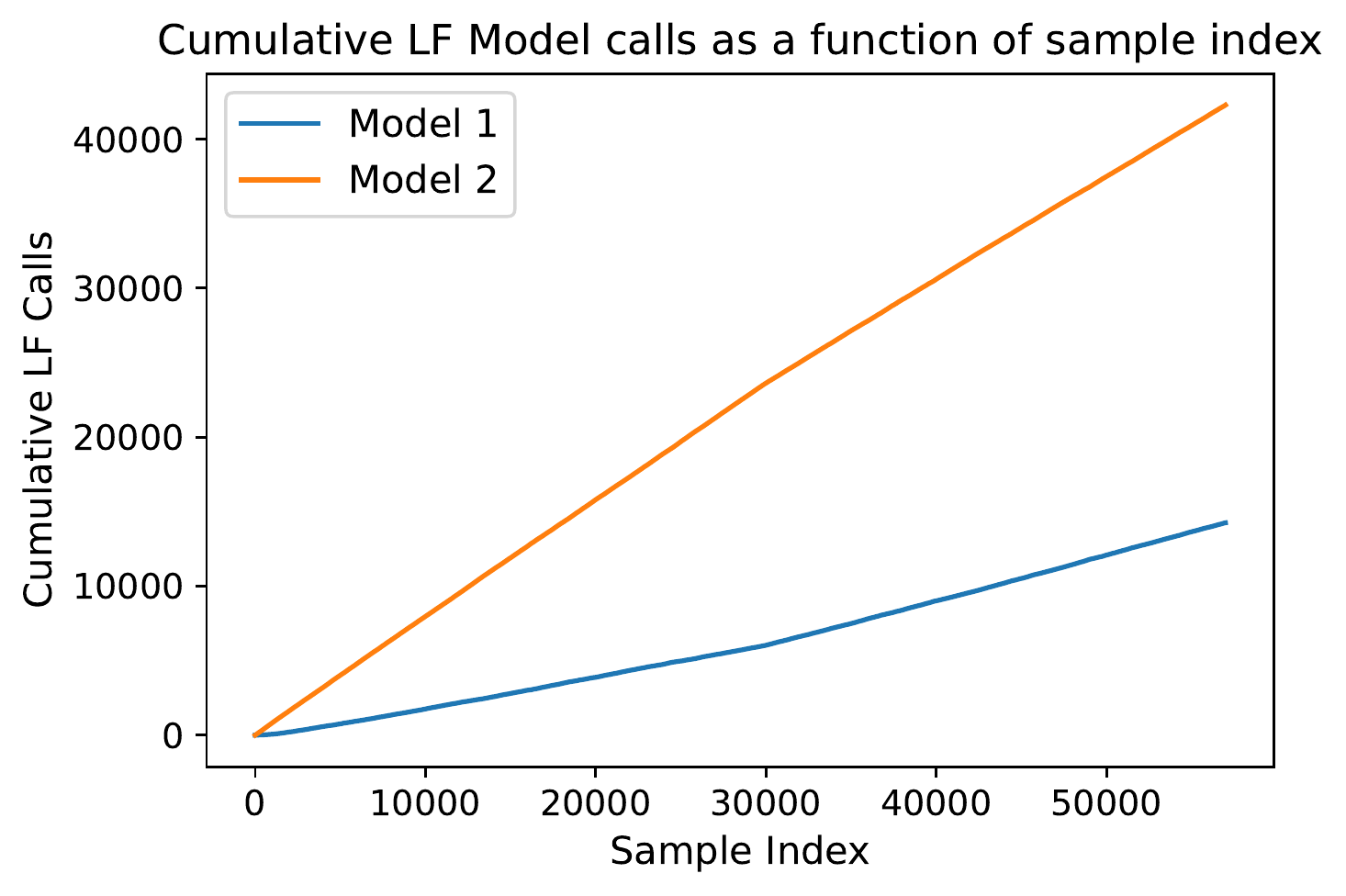}
  \caption{}
  \label{fig:Rastrigin Type 2 Cumulative LF}
\end{subfigure}
\caption{Rastrigin Function (a) Comparison between exact failure boundary and failure boundary predicted by the LFMC - LFDS surrogate for Type 1. (b) Number of model calls for each LF model using LFMC - LFDS, for Type 1 LF models. (c) Comparison between exact failure boundary and failure boundary predicted by the LFMC - LFDS surrogate for Type 2. (d) Number of model calls for each LF model using LFMC - LFDS, for Type 2 LF models.}
\label{fig:Rastrigin Function Results}
\end{figure}

Figure~\ref{fig:Rastrigin Function Results} further explores the effect of LF model selection. In both Type 1 and Type 2, the failure surface is modeled very accurately, as illustrated in Figures~\ref{fig:Rastrigin Type 1 Failure Boundaries} and~\ref{fig:Rastrigin Type 2 Failure Boundaries}. However, the allocation of samples between LF models is very different. In Type 1, where both LF models are equally good at capturing the behavior of the HF model, both LF models are called practically the same number of times (Figure~\ref{fig:Rastrigin Type 1 Cumulative LF}). But for Type 2, the cosine LF model is called much more than the quadratic LF model (Figure~\ref{fig:Rastrigin Type 2 Cumulative LF}) because it better estimates the HF model over a larger range of inputs.


\subsubsection{Discussion of Assembly Strategies and LF Model Selection}

There are many possible strategies for surrogate model assembly. We present three candidate strategies (Eq.~\eqref{eqn:Assembly_Strategies}) and the analytical examples presented thus far provide some insight into their relative performance. From Tables \ref{tab:Four-Branch_Results} and \ref{tab:Rastrigin_Results}, we see that the LFDS is typically the best strategy for minimizing computational effort, while the LFSS strategy requires slightly more HF calls. Both methods accurately estimate the failure probability. 

The LFDS and LFSS models are similar in design, except that LFDS always uses the LF model with higher probability, whereas the LFSS selects the model stochastically according to the determined model probability. Because model selection is stochastic in LFSS, the algorithm doesn't always select the ``best'' LF model. This has the effect of modestly increasing the number of necessary HF model calls. 

The LFMA case works differently than the others. It is equally accurate to the LFDS and LFSS strategies and requires a similar number of HF calls to produce a comparable failure estimate. However, it requires each LF model to be evaluated at every point, which significantly increases the total computation time due to the non-negligible runtime of the LF models themselves. Thus, the LFMA model is appreciably less efficient, requiring much more computational effort than the LFDS or LFSS strategies for a similar performance.



These examples further illustrate that the algorithm is general and will provide an accurate failure probability estimate regardless of the quality and number of LF models provided. However, the LFMC works best when each LF model has its own distinct ``domain of accuracy'', with little overlap between such domains. When each LF model captures the behavior of the HF model in a different region, the algorithm is able to pick the most accurate model for that region confidently, and the correction that needs to be applied to that model is also small and easy to learn (thus requiring fewer training points). Conversely, LF models that do not capture the behavior of the HF model require complex correction terms with large magnitudes. If another model is superior in this regime, the LFMC will ignore this model. But, if no better model can be identified, a large number of training data (i.e., HF calls) will be required. Additionally, in regions where multiple ``domains of accuracy'' overlap, the model will assign approximately equal probabilities to each model, leading to it training all surrogates in this region and increasing the number of HF calls.

\subsection{Numerical Case Studies}

In this section, we consider two more challenging numerical models. The first considers the stress-induced failure of a 3D cylinder, while the second is a practical case study of the structural failure of Tristructural isotropic-coated (TRISO) nuclear fuel. These illustrate the broad applicability of the method to problems in computational mechanics, and computational physics more generally. 

\subsubsection{Maximum von Mises Stress in a 3D Cylinder}

The first numerical example considers the stress-strain behavior of a solid 3D cylinder with stochastic elasticity parameters undergoing stochastic deformations. The cylinder has both a diameter and a height of 5 units. The bottom surface is constrained against displacement in all three directions, and the top surface is subjected to stochastic displacement in all three directions. From continuum solid mechanics, the governing equations in the absence of body forces are given by
\begin{equation}
\label{eqn:VMStressGovEqns}
    \begin{gathered}
    \Tilde{\nabla} \boldsymbol{\sigma} = 0 \\
    \boldsymbol{\sigma} = \mathbf{D} \boldsymbol{\epsilon} \\
    \Tilde{\nabla} = \begin{bmatrix}
    \frac{\partial}{\partial x_1} & 0 & 0 & \frac{\partial}{\partial x_2} & \frac{\partial}{\partial x_3} & 0 \\
    0 & \frac{\partial}{\partial x_2} & 0 & \frac{\partial}{\partial x_1} & 0 & \frac{\partial}{\partial x_3}\\
    0 & 0 & \frac{\partial}{\partial x_3} & 0 & \frac{\partial}{\partial x_1} & \frac{\partial}{\partial x_2}\\
    \end{bmatrix}
\end{gathered}
\end{equation}
where $ \boldsymbol{\sigma} $ and $ \boldsymbol{\epsilon} $ are the Cauchy stress and Cauchy strain vectors (in Voigt notation), respectively. 
The two are assumed to be linearly related through the elasticity tensor $ \mathbf{D} $. These equations are solved with the Finite Element methods (FEM) using the Tensor Mechanics module in MOOSE \cite{lindsay_moose_2022,MOOSE_MonteCarloVarianceReduction} -- an open-source, parallel, multiphysics FEM framework developed by Idaho National Laboratory and other research organizations.

The HF model treats the material as transversely isotropic such that the elasticity tensor $ \mathbf{D}_s $ is completely defined by five elasticity constants $ \{ E_x, E_z, G_{xz}, \nu_{xy}, \nu_{xz} \} $, 
and the three-dimensional domain is discretized into 11088 hexahedral elements. These constants, which are all spatially uniform, are stochastic and follow the random variables $E_1$, $E_2$, $G_3$, $\nu_1$ and $\nu_2$, whose distributions are listed in Table~\ref{tab:VMStress_Variable_List}; i.e. $ \{ E_x, E_z, G_{xz}, \nu_{xy}, \nu_{xz} \} = \{ E_1, E_2, G_3, \nu_1, \nu_2 \} $. Three LF models (denoted as LF1, LF2 and LF3) are employed, which employ a coarser mesh (1386 elements) and approximate the material as isotropic. 
The elasticity tensor for LF1 is defined by constants $ \{ E, \nu \} = \{ E_1, \nu_1 \} $, while for LF2, $ \{ E, \nu \} = \{ E_2, \nu_2 \} $. For LF3, the elasticity tensor is defined in terms of the shear modulus $ G $ given by $ G_3 $ 
and a constant Poisson's ratio $ \nu = 0.25 $. (Here, $ E $ indicates a Young's modulus, $ G $ a shear modulus, and $ \nu $ a Poisson's ratio.) In all cases, the displacements applied to the top surface of the cylinder are modeled by three random variables $ \{ U_x, U_y, U_z \} $. This is summarized in Table~\ref{tab:VMStress_Variable_List}.
\begin{table}[!ht]
    \centering
    \begin{tabular}{c c c c c c c}
    \hline \hline
      Variable Name & Distirbution & Parameters & In HF & In LF 1 & In LF 2 & In LF 3 \\
      \hline
      $ \ln (U_x), \ln (U_y), \ln (U_z) $ & Normal & $ \mu = \ln (0.15) $ & Yes & Yes & Yes & Yes \\
          &   & $ \sigma = 0.5 $ &   &   &   & \\\\
      $ \ln (E_1) $ & Normal & $ \mu = \ln (200) $ & Yes & Yes & No & No \\
          &   & $ \sigma = 0.1 $ &   &   &   & \\\\
      $ \ln (E_2) $ & Normal & $ \mu = \ln (300) $ & Yes & No & Yes & No \\
          &   & $ \sigma = 0.1 $ &   &   &   & \\\\
      $ \ln (G_3) $ & Normal & $ \mu = \ln (135) $ & Yes & No & No & Yes \\
          &   & $ \sigma = 0.1 $ &   &   &   & \\\\
      $ \ln (\nu_1) $ & Truncated Normal & $ \mu = \ln (0.15) $ & Yes & Yes & No & No \\
        &   & $ \sigma = 0.5 $ &   &   &   & \\
        &   & lower $ \to \ln (0) $ &   &   &   & \\
        &   & upper $ = \ln (0.5) $ &   &   &   & \\\\
      $ \ln (\nu_2) $ & Truncated Normal & $ \mu = \ln (0.15) $ & Yes & No & Yes & No \\
        &   & $ \sigma = 0.5 $ &   &   &   & \\
        &   & lower $ \to \ln (0) $ &   &   &   & \\
        &   & upper $ = \ln (0.5) $ &   &   &   & \\
      \hline
      \multicolumn{3}{c}{Total Number of Random Variables} & 8 & 5 & 5 & 4 \\
    \hline
    \end{tabular}
    \caption{List of Random Variables in Model Definitions for von Mises Stress Example. ($ \left\{ E_1, E_2, G_3 \right\}$ in MPa, $ \left\{ U_x, U_y, U_z \right\}$ in same units as cylinder dimensions.)}
    \label{tab:VMStress_Variable_List}
\end{table}

The response of interest is the maximum von Mises stress ($\sigma_{vm}$) in the cylinder under a given elasticity tensor and applied displacements. Failure is considered to occur when $\sigma_{vm}>90$ MPa. Given the discussion of model assembly strategies above, we apply LFDS. Table~\ref{tab:VMStress_Results} compares the LFMC - LFDS algorithm with standard SuS. Both methods use $ 5,000 $ samples per subset, and $ n_{\text{init}} = 20 $ initial training samples are used for the LFMC. The results show excellent agreement between the algorithms, with similar estimated failure probabilities and COVs. However, the LFMC requires only 373 HF model evaluations, a small fraction of those needed for standard SuS. 
\begin{table}[!ht]
    \centering
    \begin{tabular}{c c c}
    \hline \hline
         & Estimated $ P_f $ (COV) & \# HF calls (\% of total) \\
      \hline
      Subset Simulation & 3.94E-4 (0.11) & 20,000 (100\%)\\
      LFMC - LFDS & 3.98E-4 (0.12) & 373 (1.87\%) \\
    \hline
    \end{tabular}
    \caption{Results for the von Mises Stress Example (20,000 total samples generated for LFMC)}
    \label{tab:VMStress_Results}
\end{table}

\begin{figure}[!htbp]
\centering
\begin{subfigure}{.49\textwidth}
  \centering
  \includegraphics[width=\linewidth]{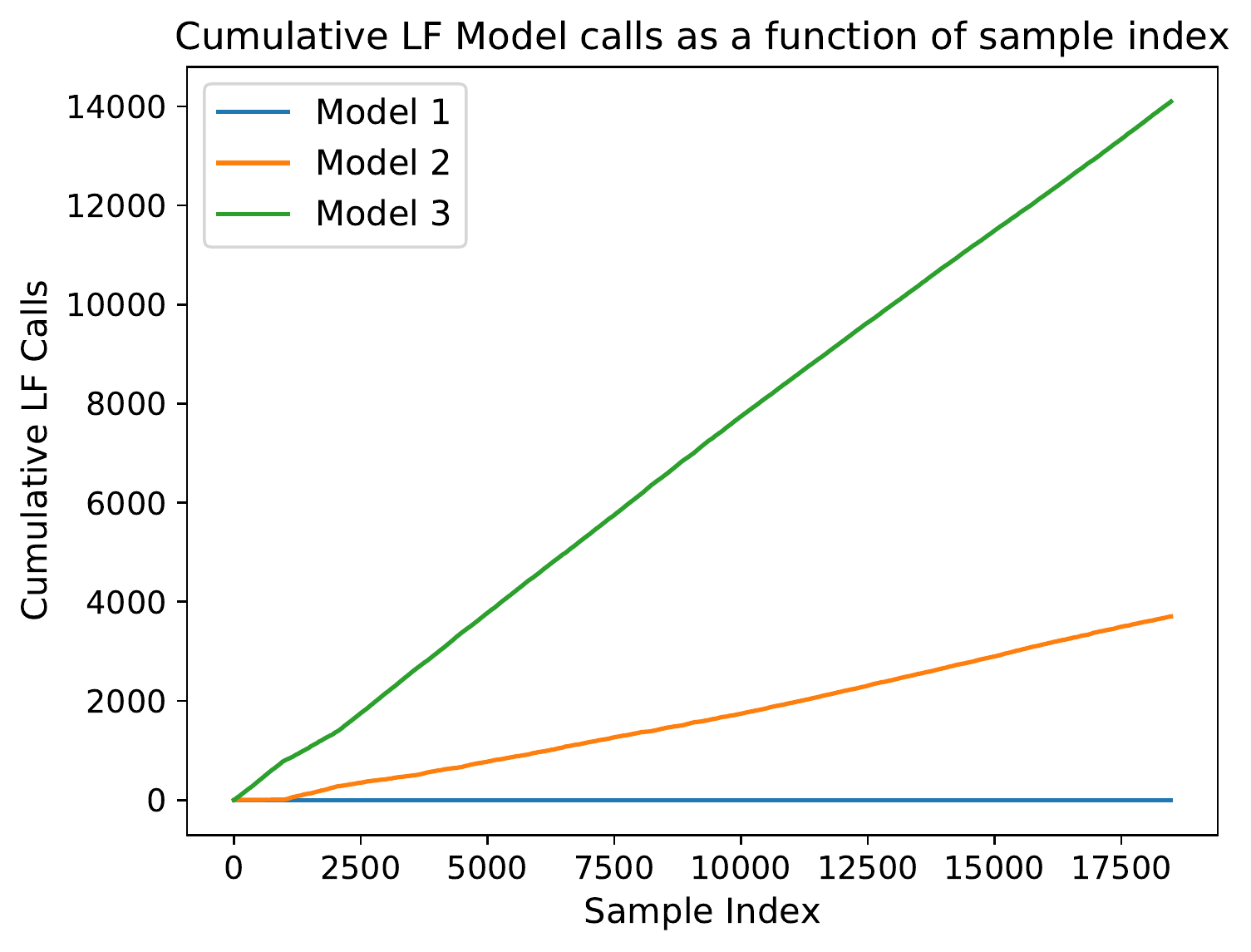}
  \caption{}
  \label{fig:VMStress Cumulative LF}
\end{subfigure}%
\begin{subfigure}{.49\textwidth}
  \centering
  \includegraphics[width=\linewidth]{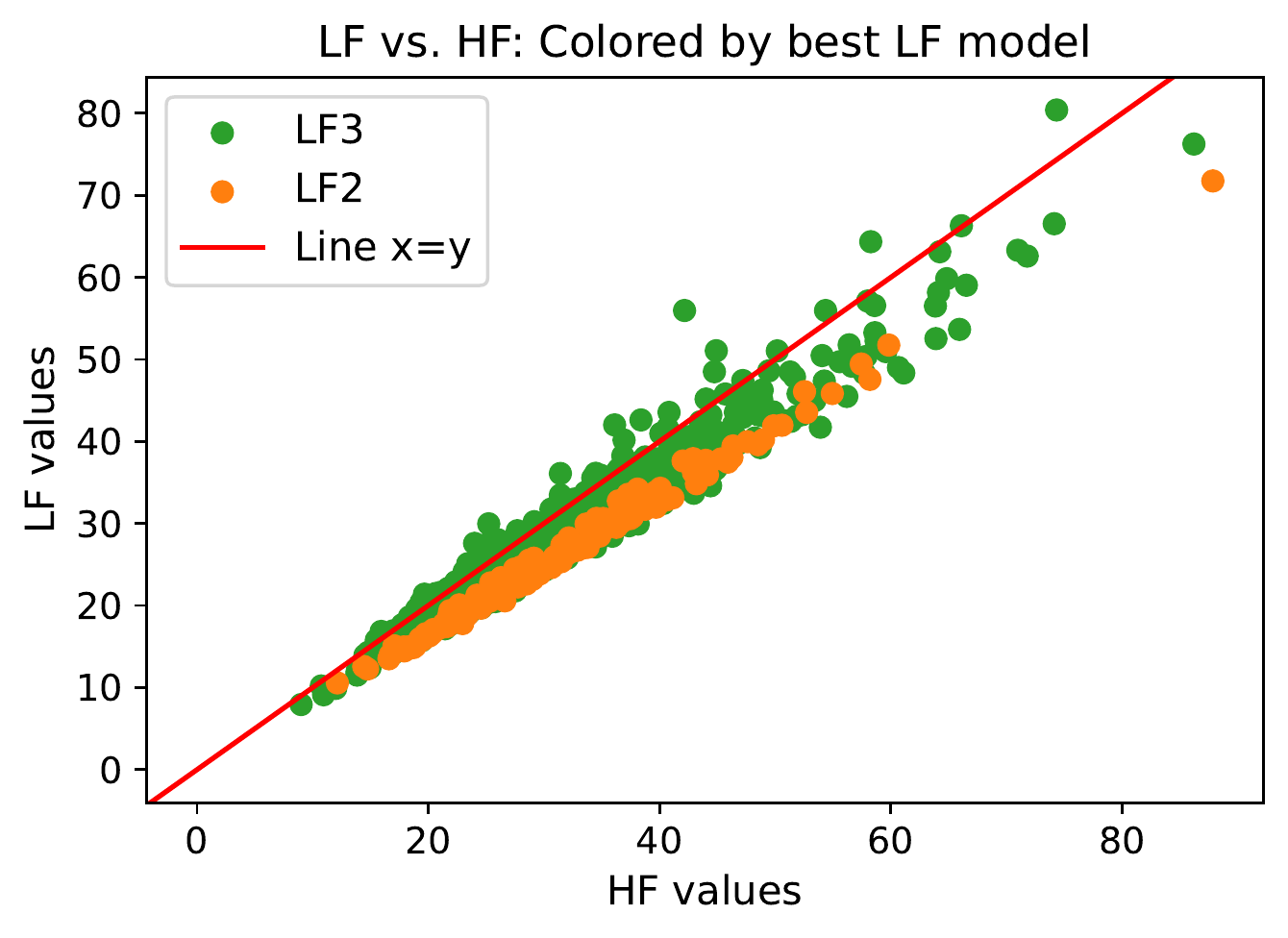}
  \caption{}
  \label{fig:VMStress Best LF Scatterplot}
\end{subfigure}
\begin{subfigure}{.33\textwidth}
  \centering
  \includegraphics[width=\linewidth]{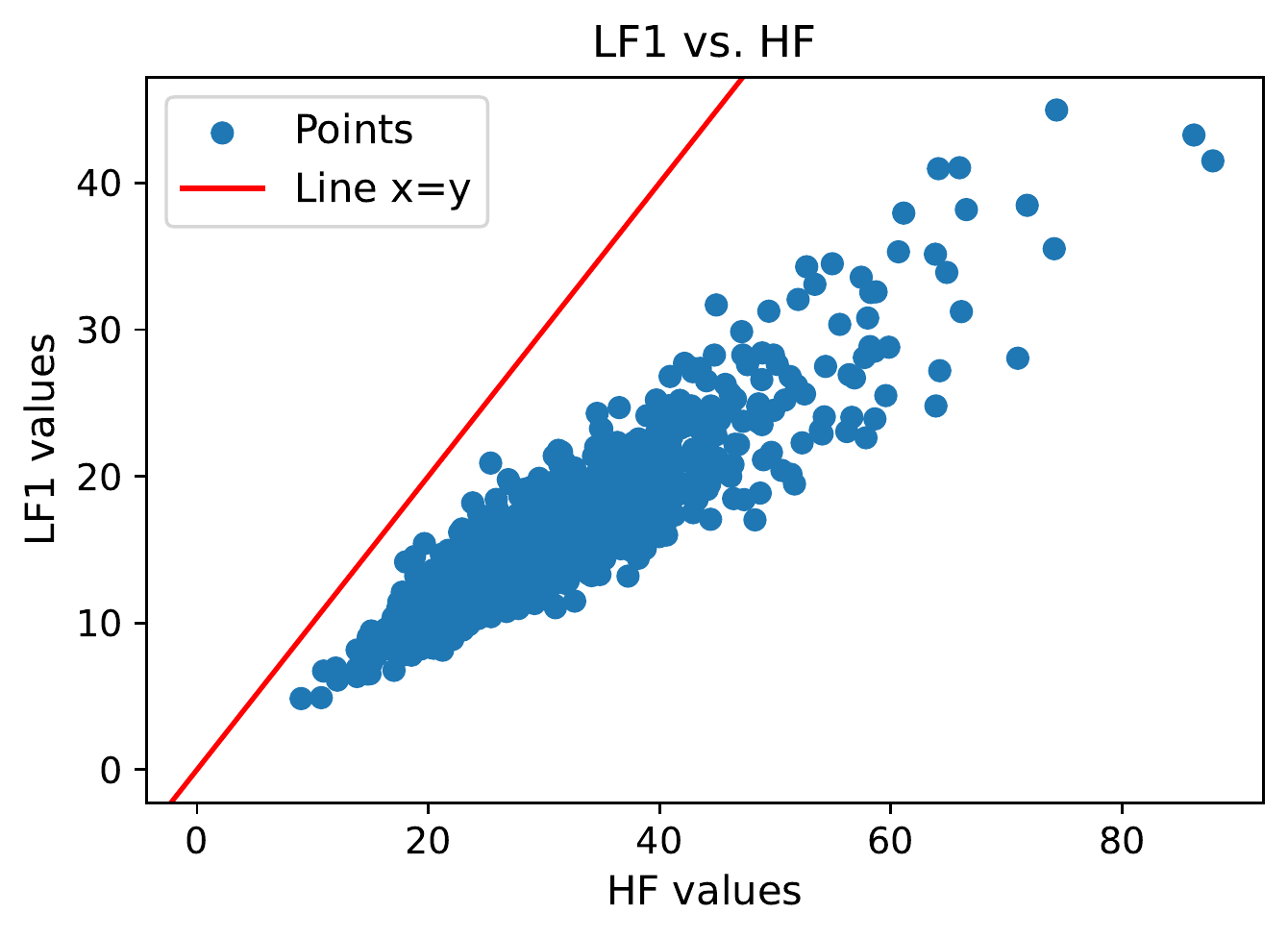}
  \caption{}
  \label{fig:VMStress LF 1 Scatterplot}
\end{subfigure}%
\begin{subfigure}{.33\textwidth}
  \centering
  \includegraphics[width=\linewidth]{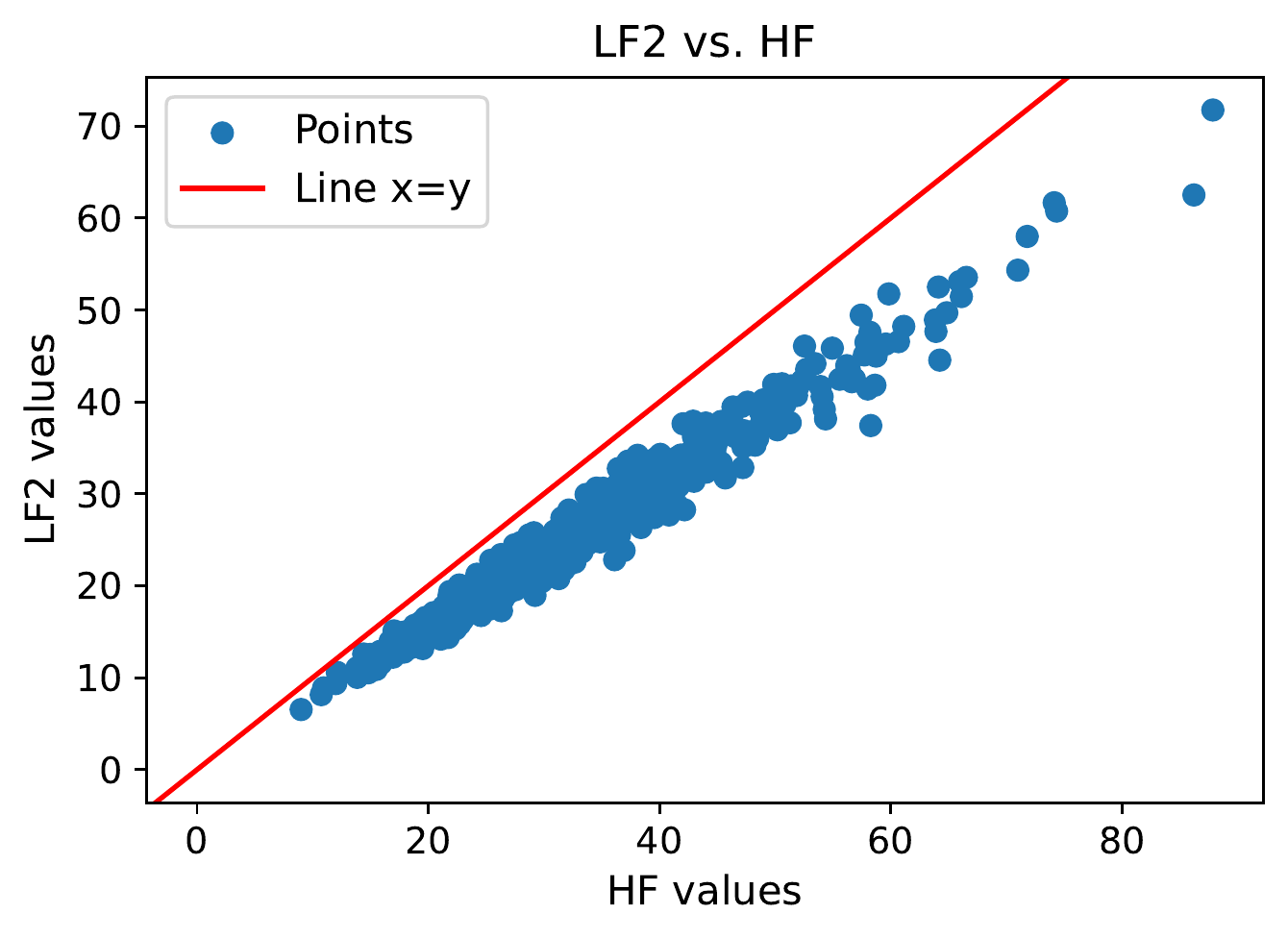}
  \caption{}
  \label{fig:VMStress LF 2 Scatterplot}
\end{subfigure}%
\begin{subfigure}{.33\textwidth}
  \centering
  \includegraphics[width=\linewidth]{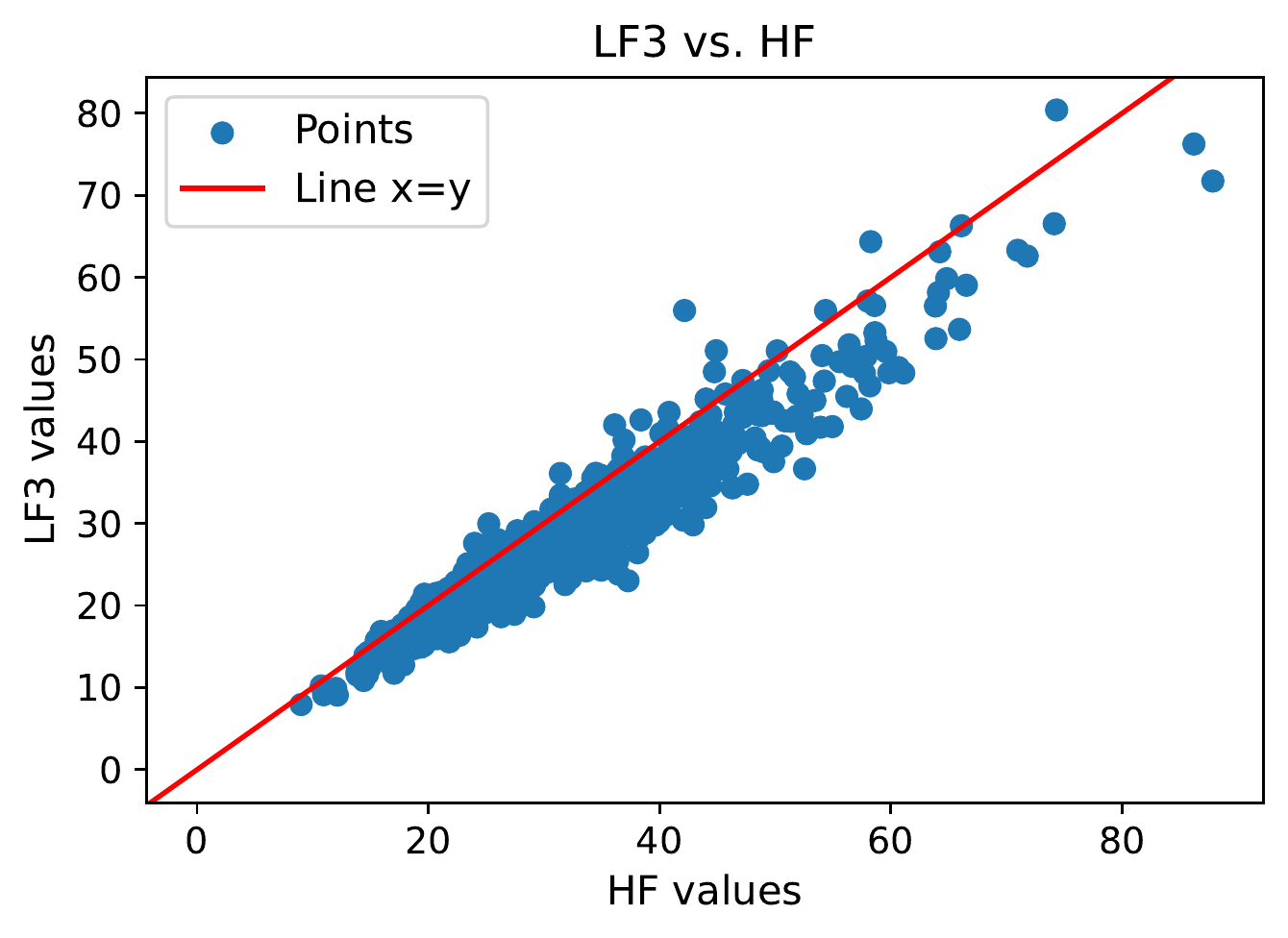}
  \caption{}
  \label{fig:VMStress LF 3 Scatterplot}
\end{subfigure}
\caption{von Mises Stress Example. (a) Number of times each LF model was called during the LFMC - LFDS algorithm. (b) Scatterplot of the best LF model prediction (uncorrected) vs. the HF response. Each point is colored according to the LF model called. (c) Scatterplot of LF1 model response vs.\ HF model response. (d) Scatterplot of LF2 model response vs.\ HF model response. (e) Scatterplot of LF3 model response vs.\ HF model response.}
\label{fig:VMStress Results}
\end{figure}

Figure~\ref{fig:VMStress Results} showcases the ability of the algorithm to illuminate the relationships between the LF models and the HF model. The graph in Figure~\ref{fig:VMStress Cumulative LF} showing the cumulative model calls for each LF model suggests that LF 3 is most often the best predictor, while LF2 is the best around 25\% of the time, and LF1 is almost never the best. To validate this, 1000 samples were drawn from the input distributions, and all four models were evaluated at each point. Figures~\ref{fig:VMStress LF 1 Scatterplot}-\ref{fig:VMStress LF 3 Scatterplot} show scatterplots for each LF model response against the HF model responses. Visual inspection of these plots corroborates the trend uncovered by the LFMC-LFDS algorithm. To further confirm this, Figure~\ref{fig:VMStress Best LF Scatterplot} gives a scatterplot showing the LF model whose prediction best matches the HF model for each point. Each point is colored according to the selected LF model. We see that most of the points are from LF3, fewer are from LF2, and none of the points are from LF1. 

For this example, all simulations were run using Sawtooth; an HPE SGI 8600-based high-performance computing system maintained by INL. Each HF model evaluation takes $ \approx 30 $ seconds, while each LF model evaluation only takes $ \approx 1.5 $ seconds, with each simulation being run on a single processor. 
Evaluating all 4 models for the $ 1000 $ Monte Carlo test samples used for Figures~\ref{fig:VMStress Best LF Scatterplot}--\ref{fig:VMStress LF 3 Scatterplot} takes $ \approx 8.33 $ hours when run serially. That number of Monte Carlo samples is far from sufficient to estimate the failure probability accurately. In contrast, the LFMC-LFDS takes only $ \approx 10 $ hours to accurately estimate the failure probability and also provides insights about model predictive quality. Note that all the LF models are of comparable cost, so selection is based on Hypothesis~\ref{hypothesis:Basic_best}, i.e., without considering cost.

\subsubsection{TRISO Fuel Model Case Study}

Tristructural isotropic-coated (TRISO) particle fuel is an advanced nuclear fuel form gaining considerable interest for advanced nuclear reactor applications \cite{DOE_TRISO}. It robustly contains fission products under high operating temperatures \cite{Hales2013,Maki2007}, which makes it an attractive fuel option for microreactors, small modular reactors, and generation-IV high-temperature reactors. The TRISO particle fuel 
is spherical and composed of four concentric layers -- the buffer layer, the inner pyrolytic carbon (IPyC) layer, the silicon carbide (SiC) layer, and the outer pyrolytic carbon (OPyC) layer -- all of which surround a fuel kernel. Maintaining the structural integrity of the coating layers of the TRISO particles during reactor operation is critical, as the SiC and PyC layers retain radioactive fission products within the particle. Failure corresponds to cracking within the layers, which may occur due to the large irradiation-induced thermo-mechanical stresses. In general, an individual particle is diagnosed to have failed when the SiC layer fails; The SiC failure is usually caused by the cracking of the IPyC layer, which causes concentrated stress in the SiC layer. Here, we focus on the case where the IPyC layer has already cracked and estimate the failure probability as the conditional probability of SiC layer cracking, i.e., $ P_f = P( \text{SiC cracking} | \text{IPyC cracked}) $, which occurs when the stress in the SiC layer exceeds its strength.

The equations governing the thermo-mechanical behavior of each layer in the TRISO particle are based on heat and momentum conservation \cite{Jiang2021,Williamson2012} summarized by
\begin{gather}
    \rho C_p \frac{\partial T}{\partial t} + \pmb{\nabla} \cdot \left( -k \pmb{\nabla} T \right) - E_f \Dot{F} = 0 \label{eqn:TRISO_heat} \\
    \pmb{\nabla} \cdot \pmb{\sigma} = 0 \label{eqn:TRISO_momentum} \\
    \pmb{\sigma} = \pmb{\mathcal{C}} : \left( \pmb{\varepsilon} - \pmb{\varepsilon_c} - \pmb{\varepsilon_t} - \pmb{\varepsilon_i} \right) \label{eqn:TRISO_constitutive}
\end{gather}
where Eq.~\ref{eqn:TRISO_heat} is the heat conservation equation, which can be used to solve for the temperature $ T $ given knowledge of the density $ \rho $, specific heat $ C_p $, thermal conductivity $ k $, energy released per fission $ E_f $, and fission rate $ \Dot{F} $. The momentum conservation equation (Eq.~\ref{eqn:TRISO_momentum}) neglects inertial and body forces, and the constitutive law (Eq.~\ref{eqn:TRISO_constitutive}) assumes small strains. Thus, the Cauchy stress $ \pmb{\sigma} $ is expressed in terms of the elasticity tensor $ \pmb{\mathcal{C}} $, the total strain $ \pmb{\varepsilon} $, the irradiation creep strain $ \pmb{\varepsilon_c} $, the thermal expansion strain $ \pmb{\varepsilon_t} $, and the irradiation-induced eigenstrain $ \pmb{\varepsilon_i} $. ($\pmb{\sigma}$, $\pmb{\mathcal{C}}$, $\pmb{\varepsilon}$, $\pmb{\varepsilon_c}$, $\pmb{\varepsilon_t}$, and $\pmb{\varepsilon_i}$ are all tensors.) These governing equations are solved numerically using the Bison fuel performance code \cite{Williamson2021}.

\begin{figure}[htbp]
\centering
\includegraphics[scale=0.495]{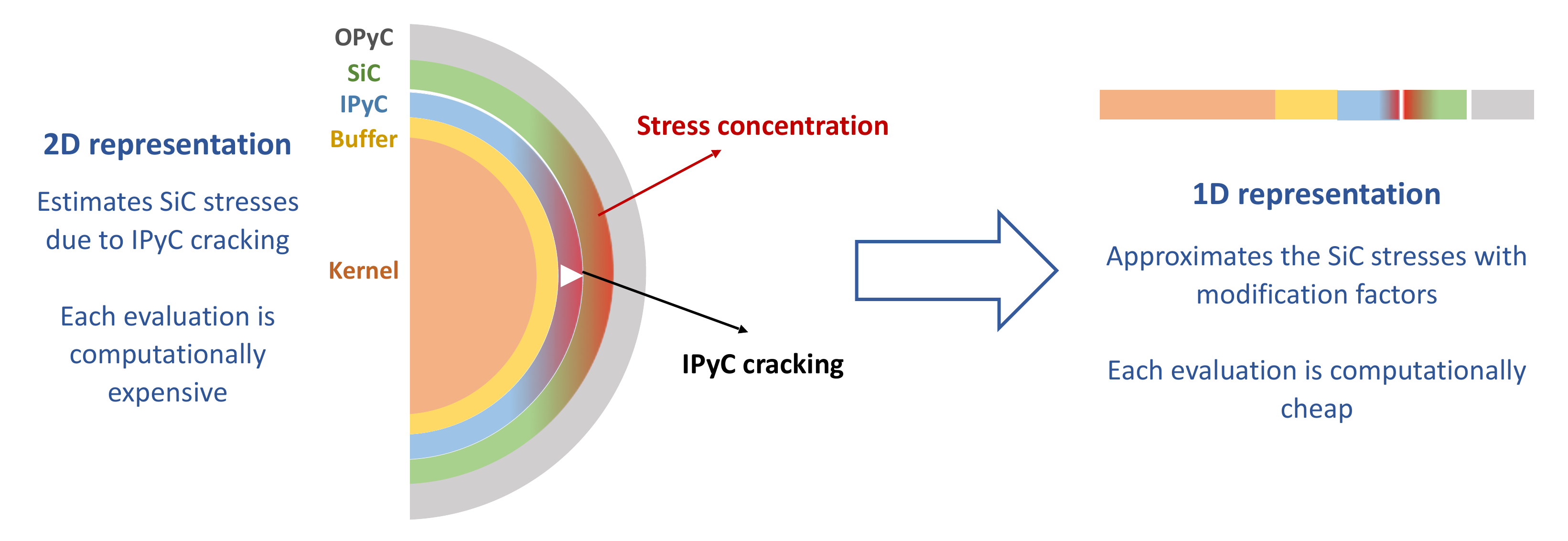}
\caption{Visualization of TRISO Numerical Models (Adapted from \cite{Dhulipala2022})}
\label{fig:TRISOModels}
\end{figure}

\begin{table}[!ht]
    \centering
    \begin{tabular}{c c c}
    \hline \hline
      Variable Name & Distribution & Parameters \\
      \hline
      Kernel Radius & Normal & $ \mu = 213.35 $ \\
        ($\mu m$) &   & $ \sigma = 4.40 $ \\\\
      Buffer Thickness & Normal & $ \mu = 98.90 $ \\
        ($\mu m$) &   & $ \sigma = 8.40 $ \\\\
      IPyC Thickness & Normal & $ \mu = 40.40 $ \\
        ($\mu m$) &   & $ \sigma = 2.50 $ \\\\
      SiC Thickness & Normal & $ \mu = 35.20 $ \\
        ($\mu m$) &   & $ \sigma = 1.20 $ \\\\
      OPyC Thickness & Normal & $ \mu = 43.40 $ \\
        ($\mu m$) &   & $ \sigma = 2.9 $ \\\\
      SiC Layer Strength ($\sigma_{SiC}^{cap}$) & Weibull & Parameters obtained \\
        (GPa) & & from \cite{Jiang2021} \\
    \hline
    \end{tabular}
    \caption{List of Random Variables in Model Definitions for TRISO Example. The variables are common to all models}
    \label{tab:TRISO_Variable_List}
\end{table}

TRISO particles are commonly numerically modeled in 1 or 2 dimensions, assuming spherical or axial symmetry, respectively, as shown in Figure~\ref{fig:TRISOModels}. The 2D model explicitly represents the crack in the IPyC layer using the extended finite element method (XFEM; \cite{Jiang2020}) and the resultant stresses in the SiC layer. The 1D model cannot explicitly represent the crack and uses stress modification factors instead to account for its effects. For our example, the HF model is a finely meshed 2D representation of the fuel particle with 6,600 elements. Two LF models are used: LF1 is a 1D representation, and LF2 is a 2D representation using a coarse mesh with 1,120 elements. In all three models, 6 random variables are used: the kernel radius, the thicknesses of the buffer, IPyC, SiC, and OPyC layers, and the strength of the SiC layer (Table~\ref{tab:TRISO_Variable_List}). Failure occurs when $ \left( \sigma_{SiC} - \sigma_{SiC}^{cap} \right) \geq 0 $, where $ \sigma_{SiC} $ is the stress in the SiC layer, and $\sigma_{SiC}^{cap}$ is the strength of the SiC layer.

Unlike the previous examples, the two LF TRISO models have vastly different computational expenses, where the 2D model is much more expensive than the 1D model. LF1 takes approximately $ 10 $ seconds on a single processor, whereas LF2 takes approximately $ 300 $ seconds on $ 4 $ processors, and the HF model takes approximately $ 900 $ seconds on $ 12 $ processors, with all models being run using Sawtooth. For this reason, we use the LFDS surrogate assembly strategy with cost-modified probability assignment. The computational complexity $ \tau $ (see section ``Incorporating Model Cost'') used here is the estimated total serial computation time for each model. 
After normalization, the computational complexity assigned to LF1, i.e., $ \tau_1 = 1 $, and that assigned to LF2, i.e., $ \tau_2 = 120 $. For numerical stability of the probability integral (Eq. \eqref{eqn:Biased_Probability}), the cost-biasing function value assigned to LF is $ \gamma \left( \tau_1 = 1 \right) = 1 $, and that assigned to LF2 is $ \gamma \left( \tau_2 = 120 \right) = 30 $; this is the ratio of the parallelized model evaluation times (i.e., ignoring the number of processors used). Back calculating from the definition, the associated biasing exponent value is found to be $ \beta \approx 0.71 $.

Table~\ref{tab:TRISO_Results} presents the failure probability results for a single LFMC - LFDC run (using $n_{\text{init}} = 20 $ initial samples) and standard SuS, both with $ 5000 $ samples per subset. 
The failure probabilities match closely, and the COV values show good agreement. However, the LFMC requires far fewer HF model calls than standard SuS. 
\begin{table}[h]
    \centering
    \begin{tabular}{c c c}
    \hline \hline
         & Estimated $ P_f $ (COV) & \# HF calls (\% of total) \\
      \hline
      Subset Simulation & 9.99E-3 (0.07) & 10000 (100\%)\\
      LFMC - LFDS & 1.17E-2 (0.09) & 58 (0.58\%) \\
    \hline
    \end{tabular}
    \caption{Results for the TRISO Example (10,000 total samples generated for LFMC)}
    \label{tab:TRISO_Results}
\end{table}

\begin{figure}[!ht]
\centering
\begin{subfigure}{.33\textwidth}
  \centering
  \includegraphics[width=\linewidth]{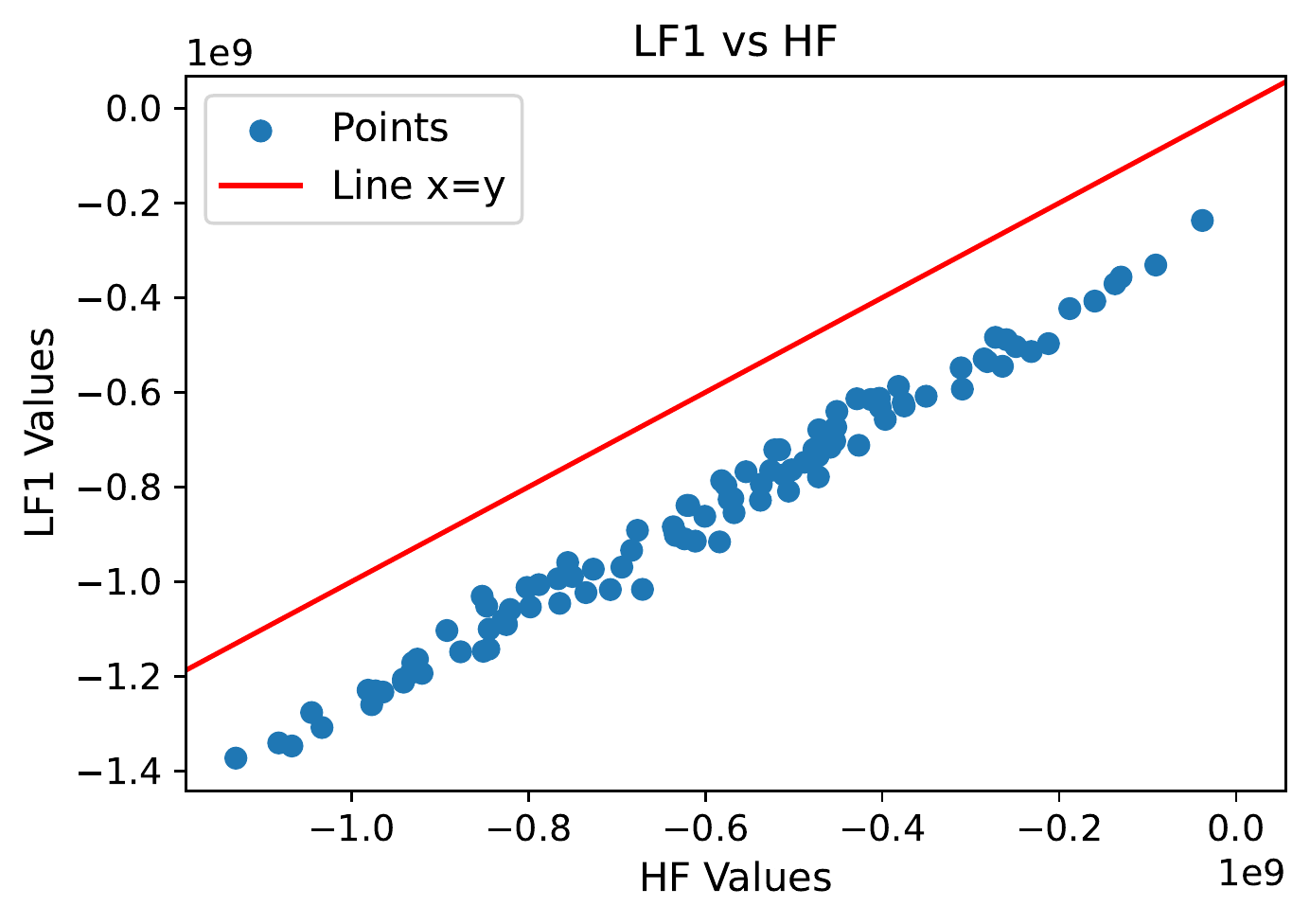}
  \caption{}
  \label{fig:TRISO LF 1 Scatterplot}
\end{subfigure}%
\begin{subfigure}{.33\textwidth}
  \centering
  \includegraphics[width=\linewidth]{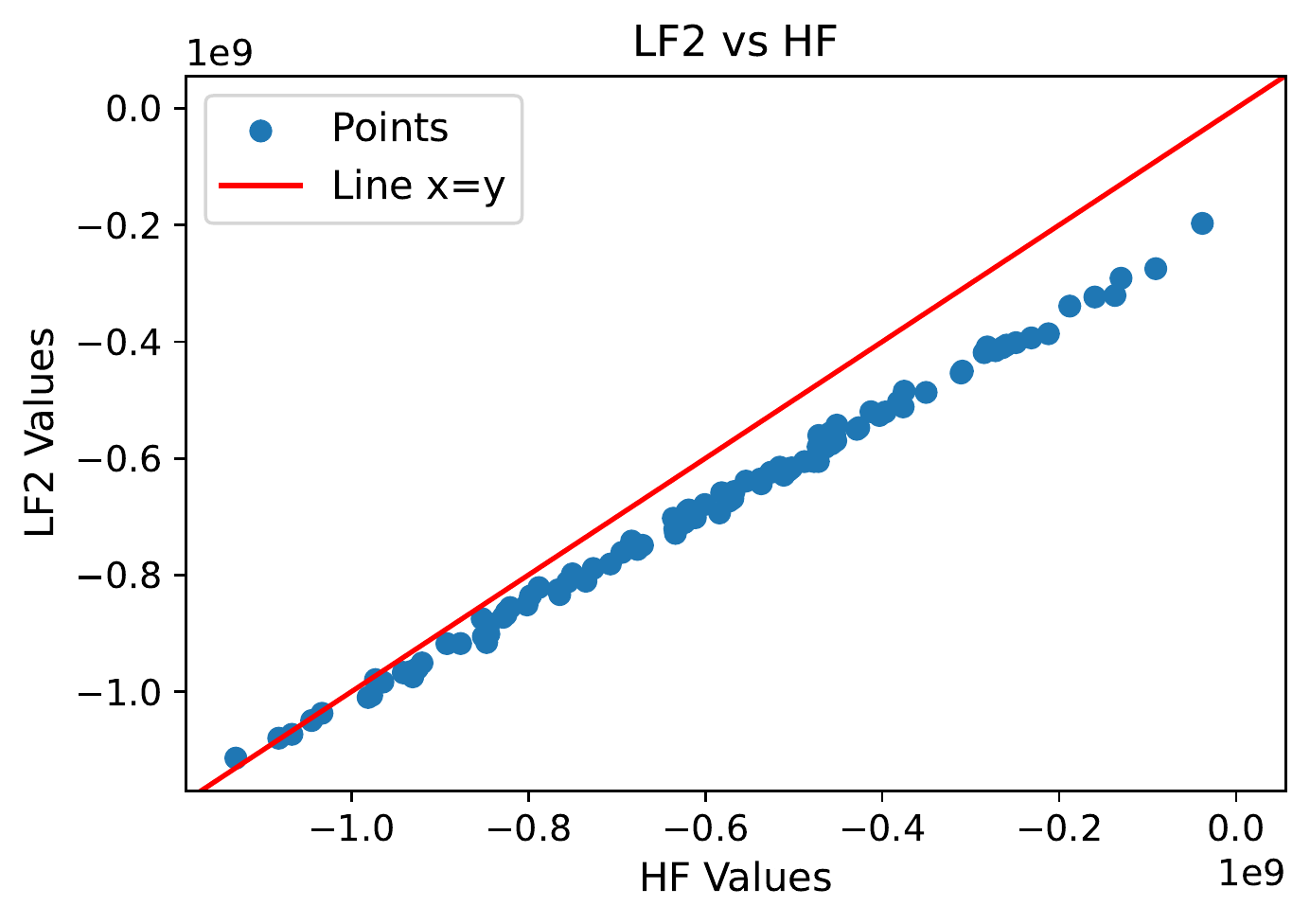}
  \caption{}
  \label{fig:TRISO LF 2 Scatterplot}
\end{subfigure}%
\begin{subfigure}{.33\textwidth}
  \centering
  \includegraphics[width=\linewidth]{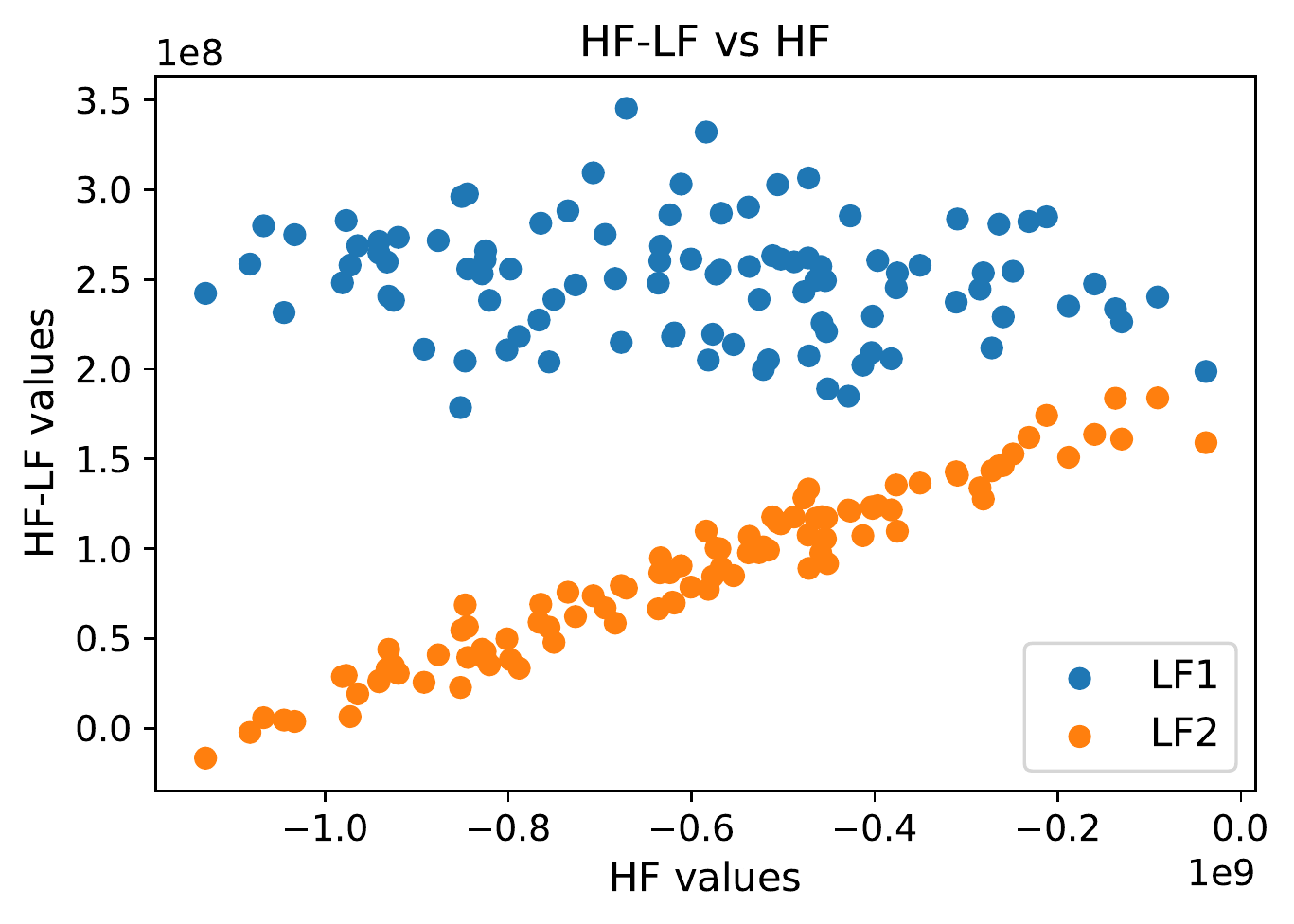}
  \caption{}
  \label{fig:TRISO HF vs HF-LF Scatterplot}
\end{subfigure}
\begin{subfigure}{.49\textwidth}
  \centering
  \includegraphics[width=\linewidth]{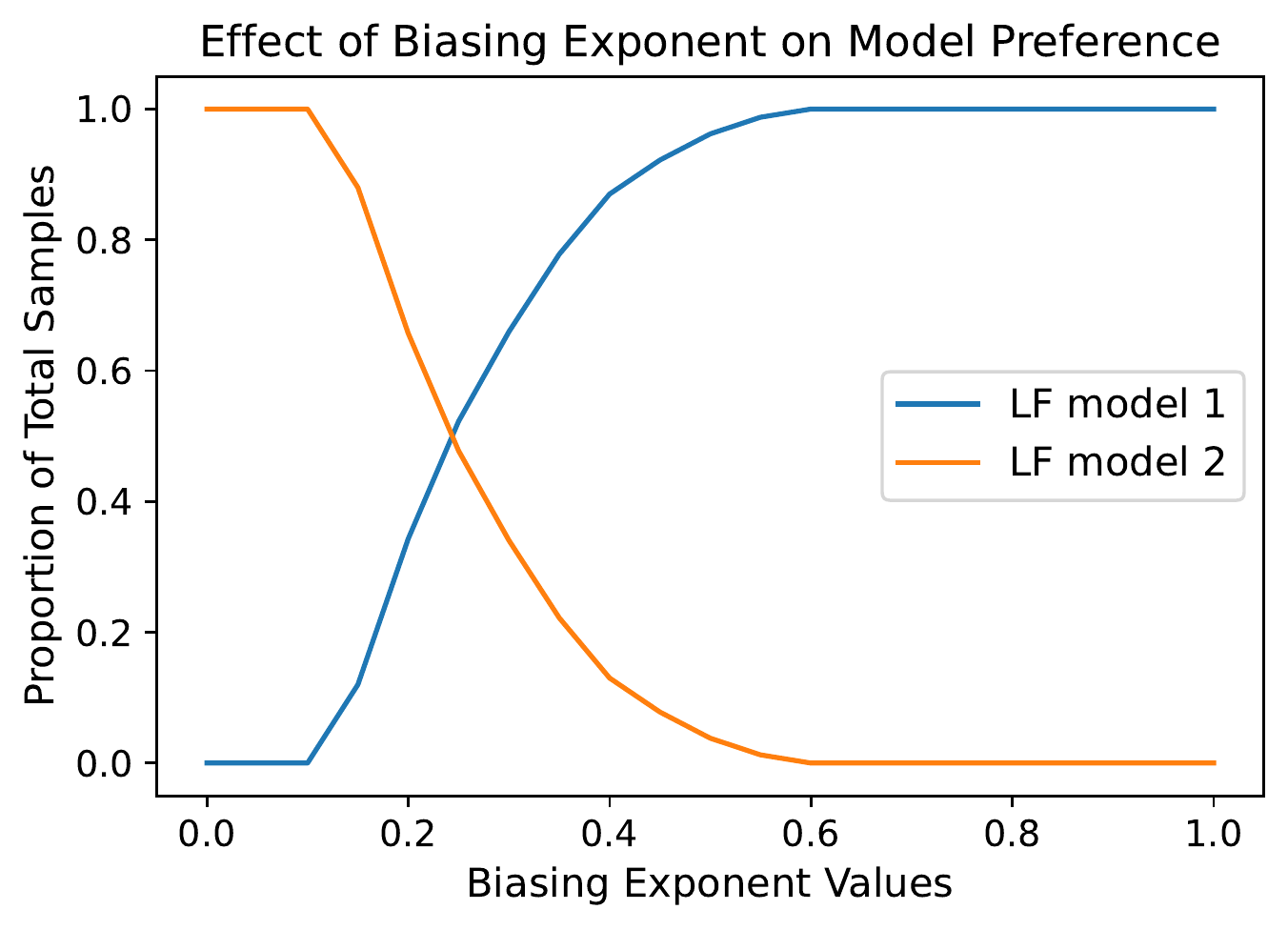}
  \caption{}
  \label{fig:TRISO Model Selection Proportions}
\end{subfigure}%
\begin{subfigure}{.49\textwidth}
  \centering
  \includegraphics[width=\linewidth]{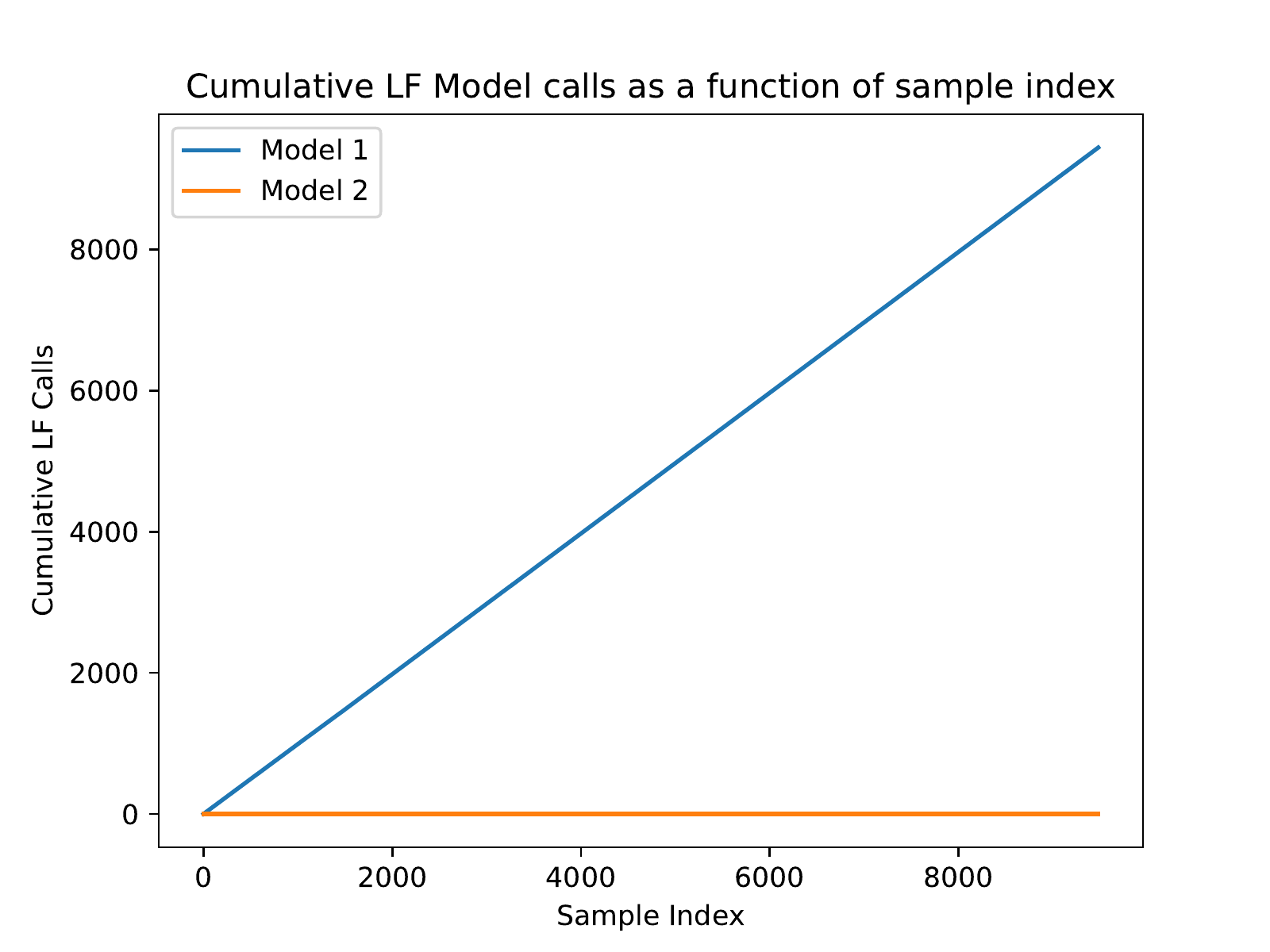}
  \caption{}
  \label{fig:TRISO Cumulative LF Calls}
\end{subfigure}
\caption{TRISO Fuel Example. (a) Scatterplot of LF1 model values vs. HF model values. (b) Scatterplot of LF2 model values vs. HF model values. (c) Scatterplot of the difference between the HF and LF model values vs. HF model values. (d) Proportion of LF1 and LF2 model calls as a function of the biasing exponent $ \beta $. (e) Cumulative number of LF1 and LF2 model calls for one run of the LFMC-LFDS algorithm with $\beta=0.71$.}
\label{fig:TRISO Results}
\end{figure}

Figure~\ref{fig:TRISO Results} further explores this performance comparison. Figures~\ref{fig:TRISO LF 1 Scatterplot} and \ref{fig:TRISO LF 2 Scatterplot} compare each LF model to the HF model and Figure~\ref{fig:TRISO HF vs HF-LF Scatterplot} plots the difference between the HF model and each LF model for 100 random samples. These plots clearly show that LF2 is consistently a better predictor than LF1. 
This, in addition to the much higher computational cost of LF2 relative to LF1, indicates that two LF models are hierarchical in nature. Given this relationship, the LFDS method based on Hypothesis~\ref{hypothesis:Basic_best} would simply prefer LF2 at all points, and LF1 would never be called, as Hypothesis~\ref{hypothesis:Basic_best} ignores the greater computational cost of LF2. By integrating cost into the model probability assignment (Hypothesis~\ref{hypothesis:Biased_best} and Eq.~\ref{eqn:Biased_Probability}), we can reduce the expense of the method. Figure~\ref{fig:TRISO Model Selection Proportions} shows the influence of $ \beta $ on LF model preference. To generate the plot, the GP correction terms associated with LF1 and LF2 were trained on a set of $ 20 $ sample points. 
Using these GPs, the cost-modified model probabilities were calculated for both models at $ 2000 $ sample points, and the ``best'' model was selected. This is repeated for several values of $\beta$ and the proportion of times each model was selected is 
plotted as a function of $ \beta $ in Figure~\ref{fig:TRISO Model Selection Proportions}. The cheaper model is clearly preferred for larger $ \beta $, while the better (but more expensive) model is preferred for lower $ \beta $. 
Figure~\ref{fig:TRISO Model Selection Proportions} suggests that LF1 should be exclusively selected to minimize overall computational effort given penalization based on parallelized model evaluation times (i.e., $ \beta \approx 0.71 $), and Figure~\ref{fig:TRISO Cumulative LF Calls} shows that this is how LFMC behaves. Note that our choice of $ \beta $ is motivated by the observed parallelized model evaluation times and is not selected as some ``optimal'' value based on the analysis presented in Figure~\ref{fig:TRISO Model Selection Proportions}.

\section{Conclusions}

One of the major bottlenecks to reliability analysis is the vast amount of computational resources and time needed to estimate failure probabilities accurately. 
This paper presents a general multi-fidelity surrogate modeling framework for efficient failure probability estimation that combines multifidelity modeling and active machine learning in an explainable, general, and robust way. 
The proposed framework uses model adaptation to construct surrogates associated with each LF model by applying a GP Regression correction term to each LF model. Each surrogate model is assigned a probability that accounts for the model's predictive accuracy, with an option to bias for its computational cost. An overall surrogate is then assembled through a model selection/averaging process based on the model probabilities. Three assembly strategies have been presented. The surrogate models are applied within subset simulation for reliability analysis, wherein an active learning filtering step is used to check surrogate model sufficiency at each simulation point. If the surrogate model is deemed insufficient, the HF model is called, and the surrogate is retrained. The procedure makes no assumptions about the type of LF models or their relationships (peer or hierarchical) with each other or with the HF. 

The framework is tested on two analytical benchmark problems and two numerical finite element case studies. The algorithm accurately estimates the failure probability in all cases while reducing the number of HF model calls necessary by $\sim2$ orders of magnitude compared to standard subset simulation. The performance of the analytical case studies highlights the algorithm's ability to learn complex failure domains. On the other hand, the finite element case studies showcased the algorithm's ability to work with complex, real-world models with moderate dimensionality, different costs, and fidelity hierarchies. 


\section*{Data Availability Statement}

The codes used to apply LFMC on the analytical examples are available on GitHub in the following repository, titled \href{https://github.com/promitchakroborty/LFMC.git}{LFMC}. All other data, models, or codes that support the findings of this study are available from the corresponding author upon reasonable request.

\section*{Acknowledgement}

This research is supported through the INL Laboratory Directed Research \& Development (LDRD) Program under DOE Idaho Operations Office Contract DE-AC07-05ID14517. This research made use of the resources of the High-Performance Computing Center at INL, which is supported by the Office of Nuclear Energy of the U.S. DOE and the Nuclear Science User Facilities under Contract No. DE-AC07-05ID14517. Portions of this work were carried out at the Advanced Research Computing at Hopkins (ARCH) core facility  (rockfish.jhu.edu), which is supported by the National Science Foundation (NSF) grant number OAC1920103.

The authors are grateful to Prof.\ Alex Gorodetsky for his feedback and suggestions on the proposed method.

\bibliography{Main}

\begin{thebibliography}{10}
\expandafter\ifx\csname url\endcsname\relax
  \def\url#1{\texttt{#1}}\fi
\expandafter\ifx\csname urlprefix\endcsname\relax\def\urlprefix{URL }\fi
\expandafter\ifx\csname href\endcsname\relax
  \def\href#1#2{#2} \def\path#1{#1}\fi

\bibitem{Zio2021}
E.~Zio, R.~B. Duffey, The risk of the electrical power grid due to natural
  hazards and recovery challenge following disasters and record floods: What
  next? (2021).
\newblock \href {http://dx.doi.org/10.1016/b978-0-12-822700-8.00008-1}
  {\path{doi:10.1016/b978-0-12-822700-8.00008-1}}.

\bibitem{Mishra2017}
S.~Mishra, O.~A. Vanli, B.~P. Alduse, S.~Jung, Hurricane loss estimation in
  wood-frame buildings using bayesian model updating: Assessing uncertainty in
  fragility and reliability analyses, Engineering Structures 135 (2017) 81--94.
\newblock \href {http://dx.doi.org/10.1016/j.engstruct.2016.12.063}
  {\path{doi:10.1016/j.engstruct.2016.12.063}}.

\bibitem{Jiang2021}
W.~Jiang, J.~D. Hales, B.~W. Spencer, B.~P. Collin, A.~E. Slaughter, S.~R.
  Novascone, A.~Toptan, K.~A. Gamble, R.~Gardner, Triso particle fuel
  performance and failure analysis with bison, Journal of Nuclear Materials
  548.
\newblock \href {http://dx.doi.org/10.1016/j.jnucmat.2021.152795}
  {\path{doi:10.1016/j.jnucmat.2021.152795}}.

\bibitem{morio2015estimation}
J.~Morio, M.~Balesdent, Estimation of rare event probabilities in complex
  aerospace and other systems: a practical approach, Woodhead publishing, 2015.

\bibitem{Schueller2004}
G.~I. Schuëller, H.~J. Pradlwarter, P.~S. Koutsourelakis, A critical appraisal
  of reliability estimation procedures for high dimensions, Probabilistic
  Engineering Mechanics 19 (2004) 463--474.
\newblock \href {http://dx.doi.org/10.1016/j.probengmech.2004.05.004}
  {\path{doi:10.1016/j.probengmech.2004.05.004}}.

\bibitem{ditlevsen1996structural}
O.~Ditlevsen, H.~O. Madsen, Structural reliability methods, Vol. 178, Wiley New
  York, 1996.

\bibitem{lemaire2013structural}
M.~Lemaire, Structural reliability, John Wiley \& Sons, 2013.

\bibitem{Rubinstein1981MonteCarlo}
R.~Y. Rubinstein, Simulation and the Monte Carlo Method, John Wiley \& Sons,
  Inc., 1981.

\bibitem{Fishman1996MonteCarlo}
G.~S. Fishman, Monte Carlo: Concepts, Algorithms, and Applications, Springer
  New York, NY, 1996.

\bibitem{MELCHERS19893}
R.~Melchers,
  \href{https://www.sciencedirect.com/science/article/pii/0167473089900039}{Importance
  sampling in structural systems}, Structural Safety 6~(1) (1989) 3--10.
\newblock \href
  {http://dx.doi.org/https://doi.org/10.1016/0167-4730(89)90003-9}
  {\path{doi:https://doi.org/10.1016/0167-4730(89)90003-9}}.
\newline\urlprefix\url{https://www.sciencedirect.com/science/article/pii/0167473089900039}

\bibitem{AU1999113}
S.~Au, C.~Papadimitriou, J.~Beck,
  \href{https://www.sciencedirect.com/science/article/pii/S0167473099000090}{Reliability
  of uncertain dynamical systems with multiple design points}, Structural
  Safety 21~(2) (1999) 113--133.
\newblock \href
  {http://dx.doi.org/https://doi.org/10.1016/S0167-4730(99)00009-0}
  {\path{doi:https://doi.org/10.1016/S0167-4730(99)00009-0}}.
\newline\urlprefix\url{https://www.sciencedirect.com/science/article/pii/S0167473099000090}

\bibitem{DERKIUREGHIAN199837}
A.~{Der Kiureghian}, T.~Dakessian,
  \href{https://www.sciencedirect.com/science/article/pii/S016747309700026X}{Multiple
  design points in first and second-order reliability}, Structural Safety
  20~(1) (1998) 37--49.
\newblock \href
  {http://dx.doi.org/https://doi.org/10.1016/S0167-4730(97)00026-X}
  {\path{doi:https://doi.org/10.1016/S0167-4730(97)00026-X}}.
\newline\urlprefix\url{https://www.sciencedirect.com/science/article/pii/S016747309700026X}

\bibitem{BUCHER1988119}
C.~G. Bucher,
  \href{https://www.sciencedirect.com/science/article/pii/0167473088900203}{Adaptive
  sampling — an iterative fast monte carlo procedure}, Structural Safety
  5~(2) (1988) 119--126.
\newblock \href
  {http://dx.doi.org/https://doi.org/10.1016/0167-4730(88)90020-3}
  {\path{doi:https://doi.org/10.1016/0167-4730(88)90020-3}}.
\newline\urlprefix\url{https://www.sciencedirect.com/science/article/pii/0167473088900203}

\bibitem{Ang1992OptimalID}
G.~L. Ang, A.~H.-S. Ang, W.~H.-C. Tang, Optimal importance‐sampling density
  estimator, Journal of Engineering Mechanics-asce 118 (1992) 1146--1163.

\bibitem{SchuellerPradlwarter1993}
G.~I. Schu\"eller, H.~J. Pradlwarter, M.~D. Pandey, {Methods for reliability
  assessment of nonlinear systems under stochastic dynamic loading - a review},
  in: EURODYN'93, 1993, pp. 751--759.

\bibitem{au2003important}
S.-K. Au, J.~Beck, Important sampling in high dimensions, Structural safety
  25~(2) (2003) 139--163.

\bibitem{AuBeck2001a}
S.~K. Au, J.~L. Beck, {Estimation of small failure probabilities in high
  dimensions by subset simulation}, Probabilistic Engineering Mechanics 16~(4)
  (2001) 263--277.
\newblock \href {http://dx.doi.org/10.1016/S0266-8920(01)00019-4}
  {\path{doi:10.1016/S0266-8920(01)00019-4}}.

\bibitem{Xiao2019}
S.~Xiao, S.~Reuschen, G.~Köse, S.~Oladyshkin, W.~Nowak, Estimation of small
  failure probabilities based on thermodynamic integration and parallel
  tempering, Mechanical Systems and Signal Processing 133.
\newblock \href {http://dx.doi.org/10.1016/j.ymssp.2019.106248}
  {\path{doi:10.1016/j.ymssp.2019.106248}}.

\bibitem{Catanach2018}
T.~A. Catanach, J.~L. Beck, \href{http://arxiv.org/abs/1804.08738}{Bayesian
  updating and uncertainty quantification using sequential tempered mcmc with
  the rank-one modified metropolis algorithm}, arXiv: Computation.
\newline\urlprefix\url{http://arxiv.org/abs/1804.08738}

\bibitem{Choi2004}
S.~K. Choi, R.~V. Grandhi, R.~A. Canfield, Structural reliability under
  non-gaussian stochastic behavior, Computers and Structures 82 (2004)
  1113--1121.
\newblock \href {http://dx.doi.org/10.1016/j.compstruc.2004.03.015}
  {\path{doi:10.1016/j.compstruc.2004.03.015}}.

\bibitem{SudretPCKriging}
R.~Schöbi, B.~Sudret, Combining polynomial chaos expansions and kriging for
  solving structural reliability problems, in: 7th Computational Stochastic
  Mechanics Conference, 2014.

\bibitem{PAPADRAKAKIS1996145}
M.~Papadrakakis, V.~Papadopoulos, N.~D. Lagaros,
  \href{https://www.sciencedirect.com/science/article/pii/0045782596010110}{Structural
  reliability analyis of elastic-plastic structures using neural networks and
  monte carlo simulation}, Computer Methods in Applied Mechanics and
  Engineering 136~(1) (1996) 145--163.
\newblock \href
  {http://dx.doi.org/https://doi.org/10.1016/0045-7825(96)01011-0}
  {\path{doi:https://doi.org/10.1016/0045-7825(96)01011-0}}.
\newline\urlprefix\url{https://www.sciencedirect.com/science/article/pii/0045782596010110}

\bibitem{HURTADO2001113}
J.~E. Hurtado, D.~A. Alvarez,
  \href{https://www.sciencedirect.com/science/article/pii/S0045782501002481}{Neural-network-based
  reliability analysis: a comparative study}, Computer Methods in Applied
  Mechanics and Engineering 191~(1) (2001) 113--132, micromechanics of Brittle
  Materials and Stochastic Analysis of Mechanical Systems.
\newblock \href
  {http://dx.doi.org/https://doi.org/10.1016/S0045-7825(01)00248-1}
  {\path{doi:https://doi.org/10.1016/S0045-7825(01)00248-1}}.
\newline\urlprefix\url{https://www.sciencedirect.com/science/article/pii/S0045782501002481}

\bibitem{Bichon2008}
B.~J. Bichon, M.~S. Eldred, L.~P. Swiler, S.~Mahadevan, J.~M. McFarland,
  Efficient global reliability analysis for nonlinear implicit performance
  functions, AIAA Journal 46 (2008) 2459--2468.
\newblock \href {http://dx.doi.org/10.2514/1.34321}
  {\path{doi:10.2514/1.34321}}.

\bibitem{AKMCS}
B.~Echard, N.~Gayton, M.~Lemaire, {AK-MCS: An active learning reliability
  method combining Kriging and Monte Carlo Simulation}, Structural Safety
  33~(2) (2011) 145--154.
\newblock \href
  {http://dx.doi.org/https://doi.org/10.1016/j.strusafe.2011.01.002}
  {\path{doi:https://doi.org/10.1016/j.strusafe.2011.01.002}}.

\bibitem{Lelievre2018a}
N.~Lelièvre, P.~Beaurepaire, C.~Mattrand, N.~Gayton, {AK-MCSi: A Kriging-based
  method to deal with small failure probabilities and time-consuming models},
  Structural Safety 73 (2018) 1--11.
\newblock \href
  {http://dx.doi.org/https://doi.org/10.1016/j.strusafe.2018.01.002}
  {\path{doi:https://doi.org/10.1016/j.strusafe.2018.01.002}}.

\bibitem{sundar2019reliability}
V.~Sundar, M.~D. Shields, Reliability analysis using adaptive kriging
  surrogates with multimodel inference, ASCE-ASME Journal of Risk and
  Uncertainty in Engineering Systems, Part A: Civil Engineering 5~(2) (2019)
  04019004.

\bibitem{Razaaly2020a}
N.~Razaaly, P.~Congedo, {Extension of AK-MCS for the efficient computation of
  very small failure probabilities}, Reliability Engineering \& System Safety
  203 (2020) 107084.
\newblock \href {http://dx.doi.org/https://doi.org/10.1016/j.ress.2020.107084}
  {\path{doi:https://doi.org/10.1016/j.ress.2020.107084}}.

\bibitem{AKSS}
X.~Huang, J.~Chen, H.~Zhu, {Assessing small failure probabilities by AK–SS:
  An active learning method combining Kriging and Subset Simulation},
  Structural Safety 59 (2016) 86--95.
\newblock \href
  {http://dx.doi.org/https://doi.org/10.1016/j.strusafe.2015.12.003}
  {\path{doi:https://doi.org/10.1016/j.strusafe.2015.12.003}}.

\bibitem{AKEE-SS}
X.~Zhang, M.~Xiao, L.~Gao, {An active learning reliability method combining
  Kriging constructed with exploration and exploitation of failure region and
  subset simulation}, Reliability Engineering \& System Safety 188 (2019)
  90--102.
\newblock \href {http://dx.doi.org/https://doi.org/10.1016/j.ress.2019.03.002}
  {\path{doi:https://doi.org/10.1016/j.ress.2019.03.002}}.

\bibitem{Xu2020}
C.~Xu, W.~Chen, J.~Ma, Y.~Shi, S.~Lu, Ak-mss: An adaptation of the ak-mcs
  method for small failure probabilities, Structural Safety 86.
\newblock \href {http://dx.doi.org/10.1016/j.strusafe.2020.101971}
  {\path{doi:10.1016/j.strusafe.2020.101971}}.

\bibitem{SingleLF}
S.~L. Dhulipala, M.~D. Shields, B.~W. Spencer, C.~Bolisetti, A.~E. Slaughter,
  V.~M. Labouré, P.~Chakroborty, Active learning with multifidelity modeling
  for efficient rare event simulation, Journal of Computational Physics 468.
\newblock \href {http://dx.doi.org/10.1016/j.jcp.2022.111506}
  {\path{doi:10.1016/j.jcp.2022.111506}}.

\bibitem{Peherstorfer2018}
B.~Peherstorfer, K.~Willcox, M.~Gunzburger, Survey of multifidelity methods in
  uncertainty propagation, inference, and optimization, SIAM Review 60 (2018)
  550--591.
\newblock \href {http://dx.doi.org/10.1137/16M1082469}
  {\path{doi:10.1137/16M1082469}}.

\bibitem{Gorodetsky2020}
A.~A. Gorodetsky, G.~Geraci, M.~S. Eldred, J.~D. Jakeman, A generalized
  approximate control variate framework for multifidelity uncertainty
  quantification, Journal of Computational Physics 408.
\newblock \href {http://dx.doi.org/10.1016/j.jcp.2020.109257}
  {\path{doi:10.1016/j.jcp.2020.109257}}.

\bibitem{pham2021ensemble}
T.~Pham, A.~A. Gorodetsky, Ensemble approximate control variate estimators:
  Applications to multi-fidelity importance sampling (2021).
\newblock \href {http://arxiv.org/abs/2101.02786} {\path{arXiv:2101.02786}}.

\bibitem{Kramer2019}
B.~Kramer, A.~N. Marques, B.~Peherstorfer, U.~Villa, K.~Willcox, Multifidelity
  probability estimation via fusion of estimators, Journal of Computational
  Physics 392 (2019) 385--402.
\newblock \href {http://dx.doi.org/10.1016/j.jcp.2019.04.071}
  {\path{doi:10.1016/j.jcp.2019.04.071}}.

\bibitem{Peherstorfer2016}
B.~Peherstorfer, T.~Cui, Y.~Marzouk, K.~Willcox, Multifidelity importance
  sampling, Computer Methods in Applied Mechanics and Engineering 300 (2016)
  490--509.
\newblock \href {http://dx.doi.org/10.1016/j.cma.2015.12.002}
  {\path{doi:10.1016/j.cma.2015.12.002}}.

\bibitem{Yang2019}
X.~Yang, D.~Barajas-Solano, G.~Tartakovsky, A.~M. Tartakovsky, Physics-informed
  cokriging: A gaussian-process-regression-based multifidelity method for
  data-model convergence, Journal of Computational Physics 395 (2019) 410--431.
\newblock \href {http://dx.doi.org/10.1016/j.jcp.2019.06.041}
  {\path{doi:10.1016/j.jcp.2019.06.041}}.

\bibitem{Yi2021}
J.~Yi, F.~Wu, Q.~Zhou, Y.~Cheng, H.~Ling, J.~Liu, An active-learning method
  based on multi-fidelity kriging model for structural reliability analysis,
  Structural and Multidisciplinary Optimization 63 (2021) 173--195.
\newblock \href {http://dx.doi.org/10.1007/s00158-020-02678-1}
  {\path{doi:10.1007/s00158-020-02678-1}}.

\bibitem{Zhang2022}
C.~Zhang, C.~Song, A.~Shafieezadeh, Adaptive reliability analysis for
  multi-fidelity models using a collective learning strategy, Structural Safety
  94.
\newblock \href {http://dx.doi.org/10.1016/j.strusafe.2021.102141}
  {\path{doi:10.1016/j.strusafe.2021.102141}}.

\bibitem{Zhang2018}
J.~Zhang, J.~Man, G.~Lin, L.~Wu, L.~Zeng, Inverse modeling of hydrologic
  systems with adaptive multifidelity markov chain monte carlo simulations,
  Water Resources Research 54 (2018) 4867--4886.
\newblock \href {http://dx.doi.org/10.1029/2018WR022658}
  {\path{doi:10.1029/2018WR022658}}.

\bibitem{GorodetskyMFNets2020}
A.~Gorodetsky, J.~D. Jakeman, G.~Geraci, M.~S. Eldred, Mfnets: Multi-fidelity
  data-driven networks for bayesian learning and prediction.

\bibitem{GorodetskyMFNets2021}
A.~A. Gorodetsky, J.~D. Jakeman, G.~Geraci, Mfnets: data efficient all-at-once
  learning of multifidelity surrogates as directed networks of information
  sources, Computational Mechanics 68 (2021) 741--758.
\newblock \href {http://dx.doi.org/10.1007/s00466-021-02042-0}
  {\path{doi:10.1007/s00466-021-02042-0}}.

\bibitem{Papaioannou2015}
I.~Papaioannou, W.~Betz, L.~Zwirglmaier, D.~Straub, {MCMC algorithms for Subset
  Simulation}, Probabilistic Engineering Mechanics 41 (2015) 89 -- 103.
\newblock \href
  {http://dx.doi.org/https://doi.org/10.1016/j.probengmech.2015.06.006}
  {\path{doi:https://doi.org/10.1016/j.probengmech.2015.06.006}}.

\bibitem{ShieldsNonGaussianSubSim}
M.~D. Shields, D.~G. Giovanis, S.~V. S., {Subset simulation for problems with
  strongly non-Gaussian, highly anisotropic, and degenerate distributions},
  Computers \& Structures 245 (2021) 106431.
\newblock \href
  {http://dx.doi.org/https://doi.org/10.1016/j.compstruc.2020.106431}
  {\path{doi:https://doi.org/10.1016/j.compstruc.2020.106431}}.

\bibitem{DhulipalaGeneralReliabilityTRISO}
S.~L.~N. Dhulipala, W.~Jiang, B.~W. Spencer, J.~D. Hales, M.~D. Shields, A.~E.
  Slaughter, Z.~M. Prince, V.~M. Labouré, C.~Bolisetti, P.~Chakroborty,
  {Accelerated statistical failure analysis of multifidelity TRISO fuel
  models}, Journal of Nuclear Materials 563.
\newblock \href
  {http://dx.doi.org/https://doi.org/10.1016/j.jnucmat.2022.153604}
  {\path{doi:https://doi.org/10.1016/j.jnucmat.2022.153604}}.

\bibitem{RasmussenWilliamsGPR}
C.~E. Rasmussen, C.~K.~I. Williams, {Gaussian Processes for Machine Learning},
  MIT Press, 2005.

\bibitem{lindsay_moose_2022}
A.~D. Lindsay, D.~R. Gaston, C.~J. Permann, J.~M. Miller, D.~Andr{\v{s}}, A.~E.
  Slaughter, F.~Kong, J.~Hansel, R.~W. Carlsen, C.~Icenhour, L.~Harbour, G.~L.
  Giudicelli, R.~H. Stogner, P.~German, J.~Badger, S.~Biswas, L.~Chapuis,
  C.~Green, J.~Hales, T.~Hu, W.~Jiang, Y.~S. Jung, C.~Matthews, Y.~Miao,
  A.~Novak, J.~W. Peterson, Z.~M. Prince, A.~Rovinelli, S.~Schunert, D.~Schwen,
  B.~W. Spencer, S.~Veeraraghavan, A.~Recuero, D.~Yushu, Y.~Wang, A.~Wilkins,
  C.~Wong, 2.0 - {MOOSE}: Enabling massively parallel multiphysics simulation,
  {SoftwareX} 20 (2022) 101202.
\newblock \href {http://dx.doi.org/10.1016/j.softx.2022.101202}
  {\path{doi:10.1016/j.softx.2022.101202}}.

\bibitem{MOOSE_MonteCarloVarianceReduction}
S.~L.~N. Dhulipala, Z.~M. Prince, A.~E. Slaughter, L.~B. Munday, W.~Jiang,
  B.~W. Spencer, J.~D. Hales, {Monte Carlo Variance Reduction in MOOSE
  Stochastic Tools Module: Accelerating the Failure Analysis of Nuclear Reactor
  Technologies}, in: International Conference on Physics of Reactors 2022
  (PHYSOR 2022), 2022, pp. 2470--2479.
\newblock \href {http://dx.doi.org/doi.org/10.13182/PHYSOR22-37608}
  {\path{doi:doi.org/10.13182/PHYSOR22-37608}}.

\bibitem{DOE_TRISO}
DOE, {TRISO Particles:} the most robust nuclear fuel on earth,
  \url{https://www.energy.gov/ne/articles/triso-particles-most-robust-nuclear-fuel-earth},
  accessed: 2021-10-24 (2019).

\bibitem{Hales2013}
J.~D. Hales, R.~L. Williamson, S.~R. Novascone, D.~M. Perez, B.~W. Spencer,
  G.~Pastore, Multidimensional multiphysics simulation of triso particle fuel,
  Journal of Nuclear Materials 443 (2013) 531--543.
\newblock \href {http://dx.doi.org/10.1016/j.jnucmat.2013.07.070}
  {\path{doi:10.1016/j.jnucmat.2013.07.070}}.

\bibitem{Maki2007}
J.~T. Maki, D.~A. Petti, D.~L. Knudson, G.~K. Miller, The challenges associated
  with high burnup, high temperature and accelerated irradiation for
  triso-coated particle fuel, Journal of Nuclear Materials 371 (2007) 270--280.
\newblock \href {http://dx.doi.org/10.1016/j.jnucmat.2007.05.019}
  {\path{doi:10.1016/j.jnucmat.2007.05.019}}.

\bibitem{Williamson2012}
R.~L. Williamson, J.~D. Hales, S.~R. Novascone, M.~R. Tonks, D.~R. Gaston,
  C.~J. Permann, D.~Andrs, R.~C. Martineau, Multidimensional multiphysics
  simulation of nuclear fuel behavior, Journal of Nuclear Materials 423 (2012)
  149--163.
\newblock \href {http://dx.doi.org/10.1016/j.jnucmat.2012.01.012}
  {\path{doi:10.1016/j.jnucmat.2012.01.012}}.

\bibitem{Williamson2021}
R.~L. Williamson, J.~D. Hales, S.~R. Novascone, G.~Pastore, K.~A. Gamble, B.~W.
  Spencer, W.~Jiang, S.~A. Pitts, A.~Casagranda, D.~Schwen, A.~X. Zabriskie,
  A.~Toptan, R.~Gardner, C.~Matthews, W.~Liu, H.~Chen, Bison: A flexible code
  for advanced simulation of the performance of multiple nuclear fuel forms,
  Nuclear Technology 207 (2021) 954--980.
\newblock \href {http://dx.doi.org/10.1080/00295450.2020.1836940}
  {\path{doi:10.1080/00295450.2020.1836940}}.

\bibitem{Dhulipala2022}
S.~L. Dhulipala, M.~D. Shields, P.~Chakroborty, W.~Jiang, B.~W. Spencer, J.~D.
  Hales, V.~M. Labouré, Z.~M. Prince, C.~Bolisetti, Y.~Che, Reliability
  estimation of an advanced nuclear fuel using coupled active learning,
  multifidelity modeling, and subset simulation, Reliability Engineering and
  System Safety 226.
\newblock \href {http://dx.doi.org/10.1016/j.ress.2022.108693}
  {\path{doi:10.1016/j.ress.2022.108693}}.

\bibitem{Jiang2020}
W.~Jiang, B.~W. Spencer, J.~E. Dolbow, Ceramic nuclear fuel fracture modeling
  with the extended finite element method, Engineering Fracture Mechanics 223.
\newblock \href {http://dx.doi.org/10.1016/j.engfracmech.2019.106713}
  {\path{doi:10.1016/j.engfracmech.2019.106713}}.

\end{thebibliography}

\end{document}